%% file: Main.tex
\documentclass[runningheads]{llncs}
\usepackage[T1]{fontenc}
\usepackage[utf8]{inputenc} 
\usepackage{hyperref}       
\usepackage{url}            
\usepackage{booktabs}       
\usepackage{amsfonts}       
\usepackage{nicefrac}       
\usepackage{microtype}      
\usepackage{xcolor}         

\input{math_commands.tex}

\usepackage{amssymb}

\usepackage{comment}

\usepackage{booktabs}
\usepackage{multirow}

\newcommand{\beginsupplement}{%
	\setcounter{table}{0}
	\renewcommand{\thetable}{A.\arabic{table}}%
	\setcounter{figure}{0}
	\renewcommand{\thefigure}{A.\arabic{figure}}%
	\setcounter{equation}{0}
	\renewcommand{\theequation}{A.\arabic{equation}}%
}

\usepackage{graphicx}
\newtheorem{assumption}{Assumption}

\newtheorem{rem}{Remark}

\newcommand{\ang}{\phi}
\newcommand{\angv}{\zeta}

\renewcommand\footnotemark{}

\begin{document}
\title{Safe Exploration Method for Reinforcement Learning under Existence of Disturbance}
\thanks{\scriptsize Preprint of a contribution submitted to the European Conference on Machine Learning and Principles and Practice of Knowledge Discovery in Databases (ECML PKDD) 2022. This preprint has not undergone  any post-submission improvements or corrections. The Version of Record of this contribution is published in M.-R. Amini et al. (Eds.): ECML PKDD 2022, LNAI 13716,  2023, and is available online at \url{https://doi.org/10.1007/978-3-031-26412-2_9}.} 
%
\titlerunning{Safe Exploration Method}
%

\author{Yoshihiro Okawa\inst{1} \and
Tomotake Sasaki\inst{1} \and
Hitoshi Yanami\inst{1} \and Toru Namerikawa\inst{2}}

\authorrunning{Y. Okawa et al.}


%


\institute{Artificial Intelligence Laboratory, Fujitsu Limited, Kawasaki, Japan \\
\email{ \{okawa.y,tomotake.sasaki,yanami\}@fujitsu.com}
\and
Department of System Design Engineering, Keio University, Yokohama, Japan\\
\email{namerikawa@keio.jp}}

\maketitle              
\begin{abstract}
\input{abst}


\end{abstract}

\section{Introduction}\label{sec:intro}
\input{Sec1}

\input{related_works}

\section{Problem formulation} \label{gen_inst}

\input{Sec2}

\section{Safe exploration method with conservative inputs} \label{headings}
\input{Sec3}

\section{Theoretical guarantee for chance constraint satisfaction} \label{theoretical_analysis}
\input{Sec4}

\section{Simulation evaluation} \label{sec:sim_pend}
\input{Sec5}

\section{Limitations}\label{sec:limit}
\input{Sec6}

\section{Conclusion}\label{conclusion}
\input{Sec7}

\subsubsection{Code Availability Statement} The source code to reproduce the results of this study is available at  
\url{https://github.com/FujitsuResearch/SafeExploration}

\subsubsection{Acknowledgements} The authors thank Yusuke Kato for the fruitful discussions on  theoretical results about the proposed method. The authors also thank anonymous reviewers for their valuable feedback. This work has been partially supported by Fujitsu Laboratories Ltd and
JSPS KAKENHI Grant Number JP22H01513.  

\bibliographystyle{splncs04}
\bibliography{Main}

\clearpage
\appendix
\beginsupplement

\section*{\centerline{\Large Appendix}}

\section{Proofs of lemmas and theorems}\label{append:proofs}
\input{append_proofs}

\section{Details of experimental setup}\label{append:ex_setup}
\input{append_sim}


\end{document}

%% file: math_commands.tex
\usepackage{amsmath,amsfonts,bm}

\def\eqref#1{equation~\ref{#1}}

\def\1{\bm{1}}

\def\vzero{{\bm{0}}}

\def\vmu{{\bm{\mu}}}
\def\vtheta{{\bm{\theta}}}
\def\vomega{{\bm{\omega}}}
\def\vvarepsilon{{\bm{\varepsilon}}}

\def\vd{{\bm{d}}}
\def\ve{{\bm{e}}}
\def\vf{{\bm{f}}}
\def\vg{{\bm{g}}}
\def\vh{{\bm{h}}}

\def\vu{{\bm{u}}}

\def\vw{{\bm{w}}}
\def\vx{{\bm{x}}}

\def\mA{{\bm{A}}}
\def\mB{{\bm{B}}}
\def\mC{{\bm{C}}}

\def\mE{{\bm{E}}}

\def\mG{{\bm{G}}}
\def\mH{{\bm{H}}}
\def\mI{{\bm{I}}}

\def\mU{{\bm{U}}}

\def\mW{{\bm{W}}}

\def\mSigma{{\bm{\Sigma}}}

\DeclareMathAlphabet{\mathsfit}{\encodingdefault}{\sfdefault}{m}{sl}
\SetMathAlphabet{\mathsfit}{bold}{\encodingdefault}{\sfdefault}{bx}{n}

\def\gN{{\mathcal{N}}}

\def\gP{{\mathcal{P}}}

\def\gX{{\mathcal{X}}}

\def\sR{{\mathbb{R}}}

\newcommand{\R}{\mathbb{R}}



%% file: abst.tex

Recent rapid developments in reinforcement learning algorithms have been giving us novel possibilities in many fields.
However, due to their exploring property, we have to take the risk into consideration
when we apply those algorithms to safety-critical problems especially in a real environment.
In this study, we deal with a safe exploration problem in reinforcement learning under the existence of disturbance. 
We define the safety during learning as satisfaction of the constraint conditions explicitly defined in terms of the state and
propose a safe exploration method that uses partial prior knowledge of a controlled object and disturbance. 
The proposed method assures the satisfaction of the explicit state constraints with a pre-specified probability 
even if the controlled object is exposed to a stochastic disturbance following a normal distribution. 
As theoretical results, 
we introduce sufficient conditions to construct conservative inputs not containing an exploring aspect used in the proposed method
and prove that the safety in the above explained sense is guaranteed with the proposed method.
Furthermore, we illustrate the validity and effectiveness of the proposed method through numerical simulations of an inverted pendulum and a four-bar parallel link robot manipulator.

%% file: Sec1.tex
Guaranteeing safety and performance during learning is one of the critical issues to implement reinforcement learning (RL) in real environments~\cite{GLAVIC20176918,9351818}. 
To address this issue, 
RL algorithms and related methods dealing with safety have been studied in recent years
and some of them are called ``safe reinforcement learning''~\cite{Garcia}.  
For example, Biyik et al.~\cite{Biyik} proposed a safe exploration algorithm 
for a deterministic Markov decision process (MDP) to be 
used in RL.
They guaranteed to prevent states from being unrecoverable by leveraging the Lipschitz continuity of its unknown transition model.
In addition, Ge et al.~\cite{Ge} proposed a 
modified Q-learning method for a constrained MDP
solved with the Lagrange multiplier method so that their algorithm 
seeks for the optimal solution ensuring that the safety premise is satisfied. 
Several methods use prior knowledge of the controlled object for guaranteeing the safety~\cite{NIPS2017_766ebcd5,liu2020safe}. 
However, few studies evaluated their safety quantitatively from a viewpoint of satisfying 
state constraints at each timestep that are defined explicitly in the problems.
Evaluating safety from this viewpoint is often useful when we have constraints on physical systems 
and need to estimate the risk caused by violating those constraints beforehand.

Recently,
Okawa et al.~\cite{Okawa} proposed a safe exploration method that is applicable to existing RL algorithms.
They quantitatively evaluated the above-mentioned safety in accordance with probabilities of satisfying the explicit state constraints.
In particular, they theoretically showed that their proposed method assures the satisfaction of the state constraints with a pre-specified probability 
by using partial prior knowledge of the controlled object such as a linear approximation model and upper bounds of the approximation errors. 
However, they did not consider the existence of external disturbance, which is an important factor when we consider safety.
	Such disturbance sometimes makes the state violate the constraints
	even if the inputs used in exploration are designed to satisfy those constraints.
Furthermore, they made a strong assumption regarding the controlled objects such that the state remains within the area satisfying the constraints 
if the input (i.e., action) is set to be zero as a conservative input that contains no exploring aspect.

In this study, we extend 
Okawa et al.'s work~\cite{Okawa} and tackle the safe exploration problem in RL under the existence of disturbance.
Our main contributions are as follows.
\begin{itemize}
	\item We  propose a novel safe exploration method for RL that uses partial prior knowledge of both the controlled object and disturbance.
	\item We introduce sufficient conditions to construct conservative inputs not containing an exploring aspect used in the proposed method. Moreover, we theoretically prove that our proposed method assures the satisfaction of explicit state constraints with a pre-specified probability under existence of disturbance following a normal distribution. 
\end{itemize}
We also demonstrate the effectiveness of the proposed method with the simulated inverted pendulum provided in OpenAI Gym~\cite{1606.01540} and a four-bar parallel link robot manipulator~\cite{namerikawa+95a} with additional disturbances. 

The rest of this paper is organized as follows:
We further compare our study with other related works in the following of this section.
In Section~\ref{gen_inst}, 
we introduce the problem formulation of this study.
In Section~\ref{headings}, we describe our safe exploration method.
Subsequently, theoretical results about  the proposed method  are shown in Section~\ref{theoretical_analysis}.
We illustrate the results of 
simulation evaluation in Section~\ref{sec:sim_pend}.
We discuss the limitations of the proposed method in Section~\ref{sec:limit}, and 
finally, we conclude this paper in Section~\ref{conclusion}.


%% file: related_works.tex
\subsection*{Comparison with related works}

Constrained Policy Optimization (CPO)~\cite{achiam2017constrained} and its extensions to solve a constrained MDP (CMDP) such as \cite{yang2020projection} are widely used to guarantee safety in RL problems.
Though those CPO-style methods do not directly evaluate their safety from a viewpoint of satisfying explicit constraints at each timestep as we discuss in our study,
they can deal with the satisfaction of (state) constraints by setting a binary constraint-violation signal as in~\cite{achiam2017constrained}. 
However, the probability that can be treated in this way is the one determined across the timesteps, 
and not the one determined at each timestep.  
Furthermore, in practice, their optimization problems often needs to be modified  and, 
in such a case, the probability depends on the solution of the modified optimization problem (the probability is specified ``post-hocly'').
In contrast, our method theoretically guarantees the satisfaction of constraints with a ``pre-specified probability'' at ``every timestep''.  
Chance constraint satisfaction at each timestep guaranteed by our method has several merits in its practical application. 
For instance, 
we can derive the probability where the constraints are satisfied in a sequence of successive timesteps.

Using initially known policy parameter that guarantees safety or pretraining with offline data are also effective. 
Chow et al.~\cite{chow2019lyapunov} proposed a safe RL algorithm based on the Lyapunov approach.
Their algorithm guarantees safety during training w.r.t the CMDP by using an initial safe policy parameter. 
Recovery RL \cite{thananjeyan2021recovery} requires pretraining to learn safety critic with offline data from some behavioral policy to guarantee safety during learning. 
In addition, Koller et al.~\cite{koller1906learning} presented a learning-based model predictive control scheme that provides high-probability safety guarantees throughout the learning process with a given controller (i.e. safe policy) that lets the states be  inside of a polytopic safe region.
As compared with these existing studies, our method requires neither pretraining or any 
policy parameter that initially guarantees the safety.

Control barrier functions (CBFs) \cite{ames2019control} also have been recently used to guarantee the safety in RL problems.
Cheng et al.~\cite{cheng2019end} showed how to modify existing RL algorithms to guarantee safety for continuous control tasks with the CBFs. 
Their method requires complete prior knowledge of the actuation dynamics in addition to partial prior knowledge of the autonomous one, 
while our method theoretically guarantees the safety with only partial prior knowledge of the autonomous and actuation dynamics of the nonlinear system.
Khojasteh et al.\cite{khojasteh2020probabilistic} proposed a learning approach for estimating posterior distribution of robot dynamics from online data to design a control policy 
that guarantees safe operation with known CBFs. 
Similarly, Fan et al.~\cite{fan2020bayesian} proposed a framework which satisfies constraints on safety, stability, and real-time performance 
while allowing the use of DNN for learning model uncertainties, whose framework leverages the theory of Control Lyapunov Functions and CBFs. 
They are different from ours since 
the former one learns the drift term and the input gain (control gain) of a nonlinear system and the latter one assumes 
the complete prior knowledge of the control gain, 
while our method uses only partial prior knowledge about the nonlinear system and does not need to learn 
the precise dynamics.

In addition, as mentioned above, some studies dealing with safety in RL require certain policies which initially guarantee their safety, but they do not provide how to design those initial policies. 
In contrast, we show sufficient conditions to construct conservative inputs that are used in our proposed method to guarantee safety during learning. This is advantageous in terms of the applicability  to real-world problems.

%% file: Sec2.tex
We consider an input-affine discrete-time nonlinear dynamic system written in the following form:
\begin{align}
\vx_{k+1}= \displaystyle \vf(\vx_k)+\displaystyle \mG(\vx_k)\vu_k+\vw_k, \label{nonlinear system}
\end{align}
where $\vx_k\in \R^n$, $\vu_k\in \R^m$, and $\vw_k\in \R^n$ stand for the state, input and disturbance at time $k$, respectively, and 
$\displaystyle \vf:\R^n\rightarrow\R^n$ and $\displaystyle \mG:\R^n\rightarrow\R^{n\times m}$ are unknown nonlinear functions. 
We suppose the state $\vx_k$ is directly observable.
An immediate cost $c_{k+1} \ge 0$ is given depending on the state, input and disturbance at each time $k$:
\begin{align}
c_{k+1}=c(\vx_k, \ \vu_k, \ \vw_k),
\end{align}
where the immediate cost function $c:\R^n\times\R^m\times\R^n \rightarrow\displaystyle[0, \infty)$ is unknown while $c_{k+1}$ is supposed to be directly observable.
We consider the situation where the constraints that the state 
is desired to satisfy from the viewpoint of safety are explicitly given by the following linear inequalities:
\begin{align}
\mH\vx\preceq \vd,  \label{state constraints} 
\end{align} 
where $\vd=[d_1,\ldots,d_{n_c}]^{\top} \in \R^{n_c}$, $\mH=[\vh_1,\ldots,\vh_{n_c}]^{\top} \in \R^{n_c\times n}$,  $n_c$ is the number of constraints and $\preceq$ means that the inequality $\leq$ holds for all elements. 
In addition, we define $\gX_s\subset\R^n$ as the set of safe states, that is, 
\begin{align}
\gX_s:=\{\vx\in \R^n|\mH\vx\preceq \vd\}.
\end{align}
Initial state $\vx_0$ is assumed to satisfy $\vx_0\in\gX_s$ for simplicity.

The primal goal of reinforcement learning is to acquire a policy (control law) that minimizes or maximizes an evaluation function with respect to the immediate cost or reward, 
using them as cues in its trial-and-error process~\cite{Sutton}.
In this study, we consider the standard discounted cumulative cost as  the evaluation function to be minimized: 
\begin{align}
J=\sum_{k=0}^{T}\gamma^k c_{k+1}. \label{evalfunc}
\end{align}
Here, $\gamma$ is a discount factor ($0<\gamma\leq1$) and $T$ is the terminal time. 

Besides (\ref{evalfunc}) for the cost evaluation, 
we define the safety in this study as satisfaction of the state constraints
and evaluate its guarantee quantitatively.
In detail, 
we consider the following chance constraint with respect to the satisfaction of the explicit state constraints (\ref{state constraints}) at each time:
\begin{align}
{\rm Pr} \{ \mH \vx_k  \preceq \vd \} \geq \eta, \label{chance constraint}
\end{align}
where ${\rm Pr} \{ \mH \vx_k  \preceq \vd \} (= {\rm Pr} \{ \vx_k \in \gX_s  \})$ denotes  the probability that $\vx_k$ satisfies the constraints (\ref{state constraints}).

The objective of the proposed safe exploration method
is to 
make 
the chance constraint (\ref{chance constraint}) 
satisfied at every time $k = 1,2, \ldots,T$ 
for a pre-specified $\eta$, where $0.5 < \eta < 1$ in this study.

\begin{figure}[hbtp]
	\begin{center}
		\includegraphics[width=.75\textwidth]{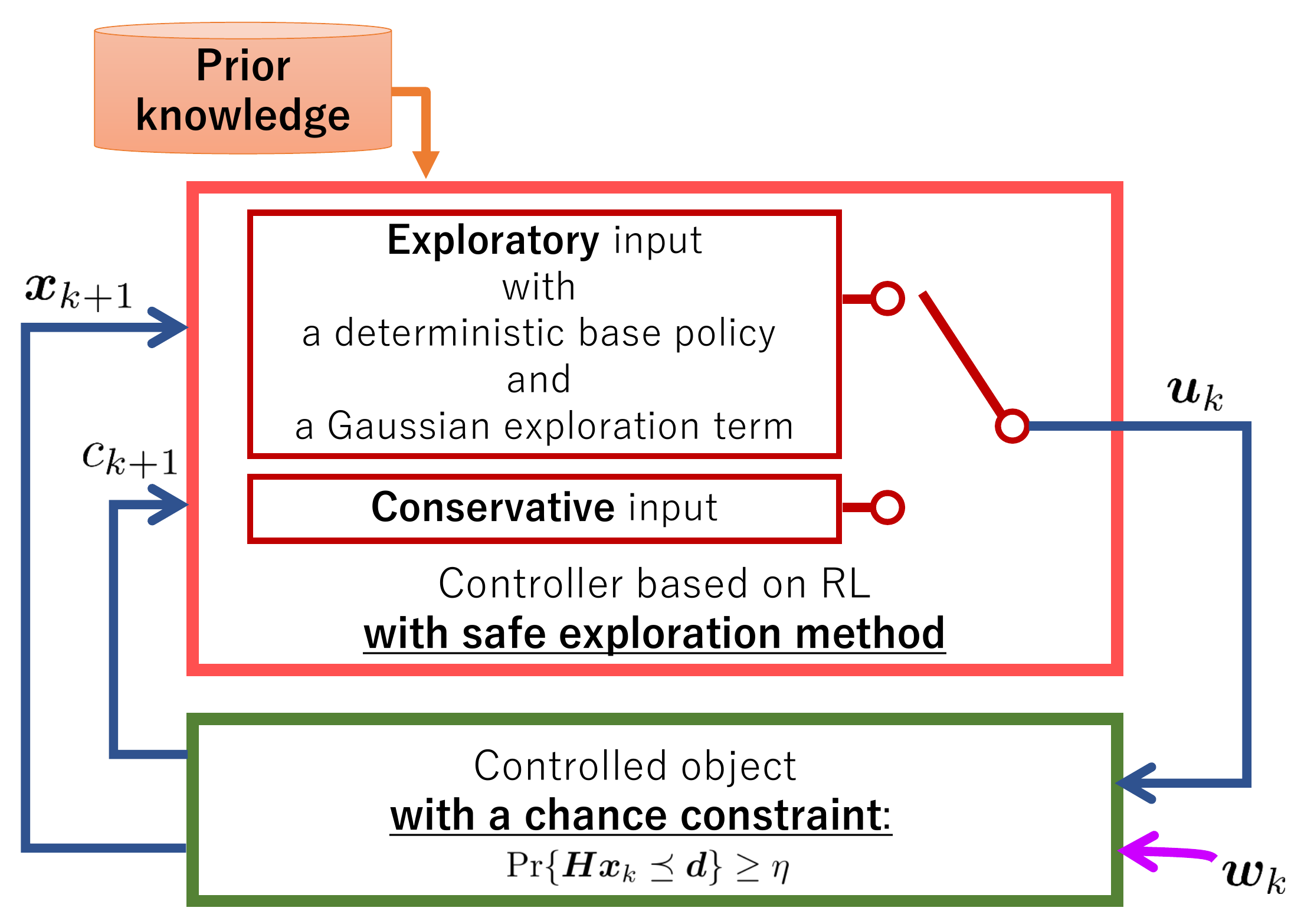}
	\end{center}
	\caption{Overview of controlled object under existence of disturbance and controller based on an RL algorithm with the proposed safe exploration method.
	The controller updates its base policy through an RL algorithm, 
	while the proposed safe exploration method makes 
	the chance constraint of controlled object satisfied by adjusting its exploration process online.}
	\label{system}
\end{figure}

Figure \ref{system} shows the overall picture of the reinforcement learning problem in this study. 
The controller in a red box generates an input $\vu_k$ according to a base policy with the proposed safe exploration method and apply it to the controlled object in a green box, which is a discrete-time nonlinear dynamic system exposed to a disturbance $\vw_k$. 
According to an RL algorithm, the base policy is updated based on the states $\vx_{k+1}$ and immediate cost $c_{k+1}$ observed from the controlled object. 
In addition to updating the base policy to minimize the evaluation function, the chance constraint should be satisfied at every time $k = 1,2, \ldots, T$.
The proposed method is described in detail in Sections~\ref{headings} and \ref{theoretical_analysis}.

As the base policy, 
we consider a nonlinear deterministic feedback control law
\begin{align}
\vmu(\ \cdot \ ; \vtheta): \sR^{n} & \to \sR^{m} \nonumber \\
                             \vx & \mapsto \vmu(\vx ; \vtheta), 
\end{align}
where $\vtheta \in \sR^{N_{\vtheta}}$ is an adjustable parameter to be updated by an RL algorithm. When we allow exploration, we generate an input $\vu_k$ by the following equation: 
\begin{align}
\vu_k = \vmu(\vx_k;\vtheta_k)+\vvarepsilon_k, \label{確率密度関数u}
\end{align}
where $\vvarepsilon_k \in\R^m$ is a stochastic exploration term 
that follows an $m$-dimensional normal distribution (Gaussian probability density function) with mean $ \vzero \in \sR^{m}$ and  variance-covariance matrix $\mSigma_k \in \sR^{m\times m}$, denoted as $\vvarepsilon_k \sim \gN(\vzero,\mSigma_k)$. In this case, as a consequence of the definition, $\vu_k$ follows a normal distribution  $\gN(\vmu_{k},\mSigma_k)$, where we define $\vmu_k:=\vmu(\vx_k;\vtheta_k)$.

We make the following four assumptions about the controlled object and the disturbance. 
The proposed method uses these prior knowledge to generate inputs, and the theoretical guarantee for chance constraint satisfactions is proven by using these assumptions.
\begin{assumption}\label{assumption:1}
	Matrices $\mA\in\R^{n\times n}$ and $\mB\in\R^{n\times m}$ in the following linear approximation model of the nonlinear dynamics (\ref{nonlinear system}) are known:
	\begin{align}
	\vx_{k+1}\simeq \mA\vx_k+\mB\vu_k+\vw_k. \label{nominal model}
	\end{align}
\end{assumption}
The next assumption is about the disturbance. 
\begin{assumption}\label{assumption:2}
		Disturbance $\vw_k$ stochastically occur according to an $n$-dimensional normal distribution $\mathcal{N} (\vmu_w, \mSigma_w)$, 
	where $\vmu_w\in\R^n$ and $\mSigma_w\in\R^{n\times n}$ are the mean and the variance-covariance matrix, respectively. The mean $\vmu_w$ and variance-covariance matrix $\mSigma_w$ are known, and the disturbance $\vw_k$ and 
	exploration term $\vvarepsilon_k$ are uncorrelated.

\end{assumption}

We define the difference 
$\ve(\vx,\vu)  \in \R^n$ 
between the nonlinear system  (\ref{nonlinear system}) and the linear approximation model (\ref{nominal model}) (i.e., approximation error) as below: 
\begin{align}
 \ve(\vx,\vu)&:=\vf(\vx)+\mG(\vx)\vu-(\mA\vx+\mB\vu). \label{approximation error} 
\end{align}
We make the following assumption on this approximation error.
\begin{assumption}\label{assumption:6}
	Regarding the approximation error $\ve(\vx,\vu)$ expressed as (\ref{approximation error}), $\bar{\delta}_j<\infty$, $\bar{\Delta}_j<\infty$,  $j=1,\ldots,n_c$ that satisfy the following inequalities are known:
	\begin{align}
	&\bar{\delta}_j \geq \sup_{\vx\in \R^n, \ \vu\in \R^m} |\vh_j^{\top} \ve(\vx,\vu)|, \ \ j=1,2,\ldots,n_c, \\
	&\bar{\Delta}_j \geq \sup_{\vx\in \R^n, \ \vu\in \R^m} |\vh_j^{\top} \left(\mA^{\tau-1}+\mA^{\tau-2}+\cdots+\mI\right) \ve(\vx,\vu)|, \  j=1,2,\ldots,n_c.
	\end{align}
\end{assumption}

The following assumption about the linear approximation model and the constraints is also made.
\begin{assumption}\label{assumption:3}
	The following condition holds for $\mB$ and $\mH=[\vh_1,\ldots,\vh_{n_c}]^{\top}$: 
	\begin{align}
		\vh_j^{\top} \mB \neq \boldsymbol{0}, \ \ \ \forall j=1,2,\ldots,n_c.
	\end{align}
\end{assumption}

Regarding the above-mentioned assumptions, Assumptions~\ref{assumption:1} and \ref{assumption:3} are similar to the assumptions used in~\cite{Okawa}, while we make a relaxed assumption on the approximation errors in Assumption~\ref{assumption:6} 
and remove assumptions on the state transition function $\vf$ and conservative inputs used in~\cite{Okawa}.

%% file: Sec3.tex
The following is the safe exploration method we propose to guarantee the safety with respect to the satisfaction of the chance constraint (\ref{chance constraint}):
{\small 
\begin{align}
	\begin{cases}
		(\mathrm{i}) \ \vu_k = \vmu(\vx_k;\vtheta_k)+\vvarepsilon_k, \ {\rm where}  \ \vvarepsilon_k \sim \mathcal{N} (\vzero, \mSigma_k) \\
		\hspace{3ex}{\rm if} \  \vx_k\!\in\!\gX_s
		\!\land\! \Big(\left\|\vh_j^{\top} \mSigma_w^\frac{1}{2}
		\right\|_2 \!\leq\! \frac{1}{\Phi^{-1}(\eta_{k}^{\prime})}(d_j\!-\!\vh_j^{\top}\hat{\vx}_{k+1}\!-\!\delta_j),  \forall \delta_j \!\in\! \{\pm \bar{\delta}_j\}, \forall j\!=\!1,\ldots,n_c \Big), \\
		(\mathrm{ii}) \ \vu_k=\vu_{k}^{stay} \\
		\hspace{3ex}{\rm if} \  \vx_k\!\in\!\gX_s \!\land\! \left(\left\|\vh_j^{\top} \mSigma_w^\frac{1}{2}
		\right\|_2 > \frac{1}{\Phi^{-1}(\eta_{k}^{\prime})}(d_j-\vh_j^{\top}\hat{\vx}_{k+1}-\delta_j), \ {\rm for \ some} \ \delta_j \in \{\pm \bar{\delta}_j\} \right), \\
		(\mathrm{iii}) \ \vu_k=\vu_k^{back} \ {\rm if} \ \vx_k\notin\gX_s,
	\end{cases}
	\label{Proposed_input}
\end{align}}\par\noindent
where $\Phi$ is the normal cumulative distribution function, 
\begin{align}
	&\hat{\vx}_{k+1}:=\mA\vx_k+\mB\vmu(\vx_k;\vtheta_k)+\vmu_w,\ \ \ \  
	\eta_{k}^{\prime}:=1-\frac{1-\left(\frac{\eta}{\xi^{k}}\right)^{\frac{1}{\tau}}}{n_c}, 
\end{align}
$\tau$ is a positive integer and $\xi$ is a positive real number that satisfies $\eta^{\frac{1}{T}}<\xi<1$. 

In the case (i), 
the variance-covariance matrix $\mSigma_k$ of $\vvarepsilon_k$ is chosen to satisfy the following inequality for all $ j=1,\ldots,n_c$:
\begin{align}
	\left\|
	\vh_j^{\top} \mB' \left[\begin{array}{cc}
		\mSigma_{k} &   \\
		& \mSigma_{w} 
	\end{array}\right]^\frac{1}{2}
	\right\|_2
	\leq\frac{1}{\Phi^{-1}(\eta_{k}^{\prime})}(d_j-\vh_j^{\top}\hat{\vx}_{k+1}-\delta_j),  \forall \delta_j \in \{\pm \bar{\delta}_j\},  
	\label{Proposed_Sigma}
\end{align}
where $\mB'=\left[\mB, \mI\right]$.

The inputs $\vu_{k}^{stay}$ and $\vu_k^{back}$ used in the  cases (ii) and (iii) are conservative inputs that are defined as follows. 
\begin{definition} \label{cond:u_tilde}
	We call $\vu_{k}^{stay}$  a conservative input of the first kind with which  ${\rm Pr}\{\mH\vx_{k+1}\preceq \vd\}\geq \left(\frac{\eta}{\xi^{k}}\right)^{\frac{1}{\tau}}$ holds if $\vx_k=\vx \in \gX_s$  occurs at time $k\geq 0$.
\end{definition}
\begin{definition} \label{cond:u_back}
	We call $\vu_k^{back}, \ \vu_{k+1}^{back}, \ \ldots, \ \vu_{k+\tau-1}^{back}$ a sequence
of conservative inputs of the second kind 
	with which for some $j\leq\tau$, ${\rm Pr}\{\vx_{k+j}\in\gX_s \}\geq \xi$  holds
if $\vx_k=\vx \notin \gX_s$ occurs at time $k\geq1$.
That is, using these inputs in this order, the state moves back to $\gX_s$ within $\tau$ steps with a probability of at least $\xi$.
%
\end{definition}
We give sufficient conditions to construct these $\vu_{k}^{stay}$ and $\vu_k^{back}$ in Section~\ref{MainResult2}.

As shown in Fig.~\ref{system} and (\ref{Proposed_input}), 
the proposed safe exploration method switches the exploratory inputs and the conservative ones 
in accordance with the current and one-step predicted state information 
by using prior knowledge of both the controlled object and disturbance,
while the previous work~\cite{Okawa} only used that of the controlled object.
In addition, 
with those prior knowledge, 
this method adjusts the degree of its exploration by restricting  the variance-covariance matrix $\mSigma_k$ of the exploration  term $\vvarepsilon_k $ to a solution of (\ref{Proposed_Sigma}).

%% file: Sec4.tex

In this section, we provide theoretical results regarding the safe exploration method we introduced in the previous section. In particular, we theoretically prove that the proposed method makes the state constraints satisfied with a pre-specified probability at every timestep. 

\begin{figure}[h]
	\begin{center}
		\includegraphics[width=1.0\textwidth]{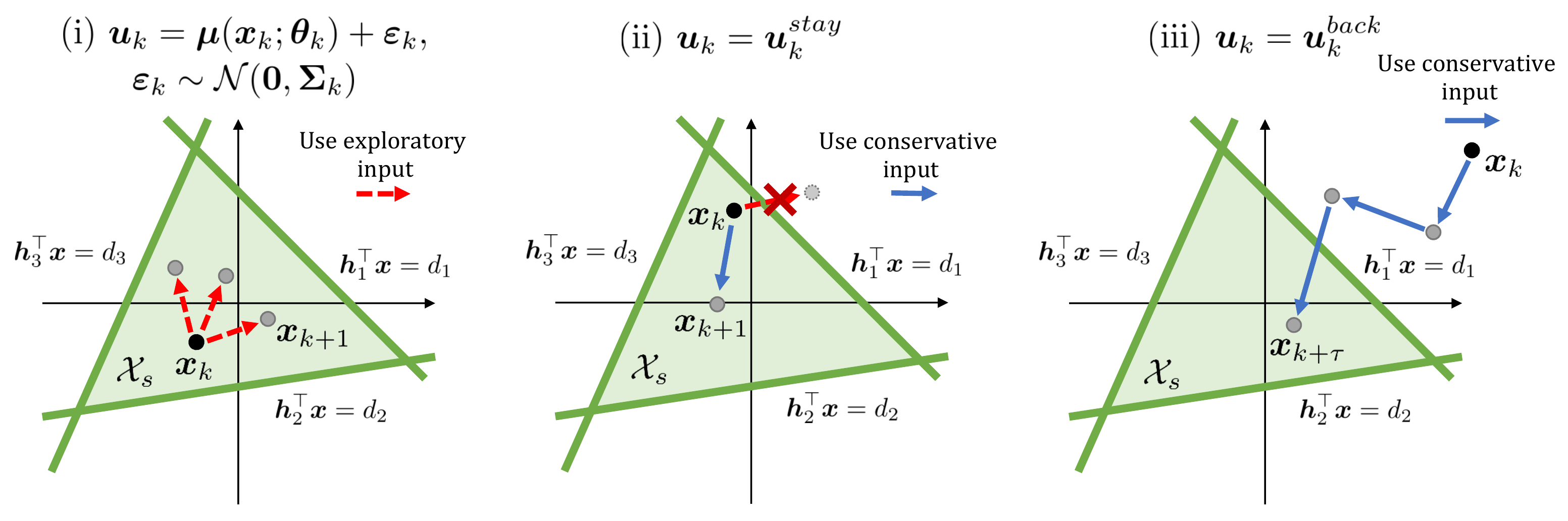}
	\end{center}
	\caption{Illustration of the proposed method for a case of $n = 2$ and $n_{c} = 3$.
	The proposed method switches two types of inputs in accordance with the current and one-step predicted state information: 
	exploratory inputs 
	generated by 
	a deterministic base policy and a Gaussian exploration term are used in the case (i), while the conservative ones that do not contain exploring aspect are used in the cases (ii) and (iii).}
	 \label{FigMethod}
\end{figure}

The proposed method (\ref{Proposed_input}) generates inputs 
differently in accordance with the following three cases (Fig.~\ref{FigMethod}): 
(i) the state constraints are satisfied and the input contains exploring aspect, (ii) the state constraints are satisfied but the input does not contain exploring aspect,  and (iii) the state constraints are not  satisfied.
We consider the case (i)  in Subsection~\ref{case_i} and the case (iii) in Subsection~\ref{case_iii}, respectively. 
We provide Theorem~\ref{Thm2} regarding the construction of conservative inputs used in the cases (ii) and (iii) in Subsection~\ref{MainResult2}. 
Then, in Subsection~\ref{MainResult1}, we provide Theorem~\ref{Thm1}, which shows  that the proposed method makes the chance constraint (\ref{chance constraint}) satisfied at every time $k$ under Assumptions~\ref{assumption:1}--\ref{assumption:3}. 
We only describe a sketch of proof of Theorem~\ref{Thm1} in this main text. 
Complete proofs of the lemmas and theorems described in this section are given in Appendix~\ref{append:proofs}.

\subsection{Theoretical result on the exploratory inputs generated with a deterministic base policy and a Gaussian exploration term} \label{case_i}

First, we consider the case when 
we generate an input containing exploring aspect 
according to (\ref{確率密度関数u}) with a deterministic base policy and a Gaussian exploration term. 
The following lemma holds.
\begin{lemma}\label{Lem2}
	Let $q\in(0.5, \ 1)$.  
Suppose Assumptions~\ref{assumption:1},~\ref{assumption:2},  and \ref{assumption:6} hold. 
	Generate input $\vu_k$ according to (\ref{確率密度関数u}) 
	when the state of the nonlinear system (\ref{nonlinear system})  at time $k$ is $\vx_k$. 
	Then, the following inequality is a sufficient condition for  ${\rm Pr}\{\vh_j^{\top}\vx_{k+1}\leq d_j\}\geq q, \ \forall j=1,\ldots,n_c$:
	\begin{align}
	\left\|
	\vh_j^{\top} \mB' \left[\begin{array}{cc}
	\mSigma_{k} &   \\
	& \mSigma_w \\
	\end{array}\right]^\frac{1}{2}
	\right\|_2
	\leq \frac{1}{\Phi^{-1}(q)}\left\{d_j-\vh_j^{\top}(\mA\vx_k+\mB\vmu(\vx_k; \vtheta_k )+\vmu_w)+\delta_j\right\},  \nonumber \\
	\forall j=1, \ 2, \ \ldots, \ n_c, \ \ \forall\delta_j\in\{\bar{\delta}_j, \ -\bar{\delta}_j\}, \label{eq:lem_explor_input}
	\end{align}
	where $\mB'=\left[\mB, \mI\right]
	$ and $\Phi$ is the normal cumulative distribution function.
\end{lemma}
Proof is given in Appendix~\ref{append:proof_Lem2}.
This lemma is proved with 
the equivalent transformation of the chance constraints into their deterministic counterparts~\cite{Boyd} 
and holds since the disturbance $\vw_k$ follows a normal distribution and is uncorrelated to the input $\vu_{k}$ according to Assumption~\ref{assumption:2} and (\ref{確率密度関数u}).	
Furthermore, this lemma shows that, in the case (i), the state satisfies the constraints  with an arbitrary probability $q\in(0.5,1)$ by adjusting the variance-covariance matrix $\mSigma_{k}$
used to generate the Gaussian exploration term $\vvarepsilon_{k}$ 
so that the inequality (\ref{eq:lem_explor_input}) would be satisfied.

\subsection{Theoretical result on the conservative inputs of the second kind} \label{case_iii}
Next, we consider the case when the state constraints are not satisfied. In this case, we use the conservative inputs defined in Definition~\ref{cond:u_back}. 
Regarding this situation, the following lemma holds.
\begin{lemma}\label{Lem1}
	Suppose we use input sequence $\vu_k^{back}, \ \vu_{k+1}^{back}, \ \ldots, \ \vu_{k+j-1}^{back}$ ($j < \tau$) given in Definition~\ref{cond:u_back}   
	when $\vx_{k-1} \in \gX_s$ and $\vx_k=\vx \notin \gX_s$ occur. 
	Also  suppose $\vx_k\in\gX_s \Rightarrow {\rm Pr}\{\vx_{k+1}\in\gX_s\}\geq p $ holds with $p \in (0,1)$. 
	Then ${\rm Pr}\{\vx_{k}\in\gX_s\}\geq \xi^k p^{\tau}$ holds for all $k = 1, 2, \dots, T$ if $\vx_0\in\gX_s$. 
\end{lemma}
Proof is given in Appendix~\ref{append:proof_Lem1}.
This lemma gives us a theoretical guarantee to make a state violating the constraints  satisfy them with a desired probability after a certain number of timesteps if we use conservative inputs (or input sequence) defined in Definition~\ref{cond:u_back}.


\subsection{Theoretical result on how to generate conservative inputs} \label{MainResult2}

As shown in (\ref{Proposed_input}), our proposed method uses conservative inputs $\vu_{k}^{stay}$ and $\vu_k^{back}$ given in  Definitions~\ref{cond:u_tilde} and \ref{cond:u_back}, respectively.
Therefore, when we try to apply this method to real problems, we need to 
construct such conservative inputs.
To address this issue, in this subsection, we introduce sufficient conditions to construct those conservative inputs, which are given by using prior knowledge of the controlled object and disturbance. Namely, regarding  $\vu_{k}^{stay}$ and $\vu_k^{back}$ used in (\ref{Proposed_input}), we have the following theorem. 

\begin{theorem}\label{Thm2}
	Let $q \in (0.5, 1)$. 
	Suppose Assumptions~\ref{assumption:1},~\ref{assumption:2} and \ref{assumption:6} hold. 
Then, if input $\vu_k$ satisfies the following inequality for all $j=1, 2, \ldots, n_c$ and $\delta_j \in\{\bar{\delta}_j, \ -\bar{\delta}_j\}$,   ${\rm Pr}\{\vx_{k+1}\in\gX_s\}\geq q$ holds:
	\begin{align}
	d_j - \vh_j^{\top}(\mA\vx_k + \mB\vu_k + \vmu_w) - \delta_j  \geq \Phi^{-1}(q')\left\| \vh_j^{\top} \mSigma^{\frac{1}{2}}_w \right\|_2, 
	\label{theorem3}
	\end{align}
where $q'=1-\frac{1-q}{n_c}$. 

	In addition, if input sequence 
	$\mU_k = [\vu_k^{\top},  \vu_{k+1}^{\top},  \ldots,  \vu_{k+\tau-1}^{\top}]^{\top}$ satisfies the following inequality for all $j=1, 2, \ldots, n_c$ and $\Delta_j\in \{  -  \bar{\Delta}_j, \bar{\Delta}_j \}$, ${\rm Pr}\{\vx_{k+\tau}\in\gX_s\}\geq q $ holds:
	{\small 
	\begin{align} 
	d_j - \vh_j^{\top}\left(\mA^\tau \vx_k + \hat{\mB} \mU_k + \hat{\mC} \hat{\vmu}_w \right) - \Delta_j \geq 
	\Phi^{-1}(q')\left\| \vh_j^{\top} \hat{\mC}
	\left[
	\begin{array}{ccc}
	\mSigma_w & &  \\
	& \ddots & \\
	& & \mSigma_w
	\end{array}
	\right]^\frac{1}{2}\right\|_2, 
	\label{theorem2}
	\end{align}}\par\noindent
	where
	$\hat{\vmu}_w=\left[\vmu_w^{\top}, \ldots, \vmu_w^{\top} \right]^{\top} \in \mathbb{R}^{n\tau}$, 
	$\hat{\mB} = [\mA^{\tau-1}\mB,  \mA^{\tau-2}\mB, \ldots, \mB] $ and 
	$\hat{\mC} = [\mA^{\tau-1},  \mA^{\tau-2},  \ldots,  \mI]$. 
\end{theorem}
Proof is given in Appendix~\ref{append:proof_Thm2}.
This theorem means that, if we find solutions of (\ref{theorem3}) and (\ref{theorem2}), 
they can be used as the conservative inputs $\vu_{k}^{stay}$ and $\vu_k^{back}$.  
Since (\ref{theorem3}) and (\ref{theorem2}) are linear w.r.t. $\vu_k$ and $\mU_k$, we can use solvers for linear programming to find solutions. 
Concrete examples of the the conditions given in this theorem are shown in our simulation evaluations in Section~\ref{sec:sim_pend}. 

\subsection{Main theoretical result: Theoretical guarantee for chance constraint satisfaction} \label{MainResult1}

Using the complementary theoretical results described so far,  
we show our main theorem that guarantees the satisfaction of the safety when we use our proposed safe exploration method (\ref{Proposed_input}), even with the existence of disturbance.


\begin{theorem} \label{Thm1}
	Let $\eta \in (0.5, 1)$.  
	Suppose Assumptions~\ref{assumption:1} through \ref{assumption:3} hold. 
	Then, by determining input $\vu_{k}$ according to the proposed method (\ref{Proposed_input}), 
	chance constraints (\ref{chance constraint}) are satisfied at every time $k=1,2, \ldots,T$.
\end{theorem}
\textit{Sketch of Proof.}~
First, consider the case of $(\mathrm{i})$ in (\ref{Proposed_input}). 
From Lemma~\ref{Lem2}, Assumptions~\ref{assumption:6} and \ref{assumption:3}, and  Bonferroni's inequality,
\begin{align}
	{\rm Pr}\{\mH\vx_{k+1}\preceq \vd\}\geq\left(\frac{\eta}{\xi^{k}}\right)^{\frac{1}{\tau}} \label{機会制約１}
\end{align}
holds if the input $\vu_k$ is determined by  (\ref{確率密度関数u})  with $\mSigma_{k}$ satisfying (\ref{Proposed_Sigma}),  
and thus, 
chance constraints (\ref{chance constraint}) are satisfied for $k = 1,2, \dots, T$.

Next, in the case of $(\mathrm{ii})$ in (\ref{Proposed_input}), by determining an input as $\vu_k=\vu_{k}^{stay}$ that is defined in Definition~\ref{cond:u_tilde}, ${\rm Pr}\{\mH\vx_{k+1}\preceq \vd\}\geq\left(\frac{\eta}{\xi^{k}}\right)^{\frac{1}{\tau}}$ holds  when $\vx_k\in\gX_s$.

Finally, by determining input as $\vu_k=\vu_k^{back}$ in case $(\mathrm{iii})$ of (\ref{Proposed_input}), ${\rm Pr}\{\mH\vx_k\preceq \vd\}\geq \eta$ holds for any $\vx_k\in\R^n$, $k=1,2,\ldots,T$ from Lemma~\ref{Lem1}.
Hence, noting $\left(\frac{\eta}{\xi^{k}}\right)^{\frac{1}{\tau}} > \eta$, ${\rm Pr}\{\mH\vx_k\preceq \vd\}\geq \eta$ is satisfied for $k=1,2, \ldots, T$.
Full proof is given in Appendix~\ref{append:proof_Thm1}.
\qed

The theoretical guarantee of safety proved in Theorem~\ref{Thm1}  is obtained with the equivalent transformation of the chance constraints into their deterministic counterparts under the assumption on disturbances (Assumption~\ref{assumption:3}).
That is,  this theoretical result holds 
since the disturbance follows a normal distribution and is uncorrelated to the input.	
The proposed method, however,  can be applicable to deal with other types of disturbance
if the sufficient part holds with a certain transformation. \linebreak

%% file: Sec5.tex
\subsection{Simulation conditions}\label{subsec:sim_con}
We evaluated the validity of the proposed method with an inverted-pendulum problem provided as ``Pendulum-v0'' in OpenAI Gym~\cite{1606.01540} and a four-bar parallel link  robot manipulator with two degrees of freedom dealt in \cite{namerikawa+95a}.
Configuration figures of both problems are illustrated in Fig.~\ref{fig:sim_fig} in Appendix~\ref{append:fig}.
We added external disturbances to these problems.

\subsubsection{Inverted-pendulum:}
A discrete-time 
dynamics of this problem is given by
\begin{align}
	\left[
	\begin{array}{c}
		\ang_{k+1} \\
		\angv_{k+1}
	\end{array}
	\right] 
	= 
	\left[
	\begin{array}{c}
		\ang_k + T_s\angv_k \\
		\angv_k -T_s \frac{3g}{2\ell}\sin(\ang_k+\pi)   
	\end{array}
	\right] +
	\left[
	\begin{array}{c}
		0 \\
		T_s\frac{3}{m\ell^2}
	\end{array}
	\right] u_k + \vw_k,\label{eq:sim_real}
\end{align}
where 
$\ang_k\in\R$ and 
$\angv_k\in\R$ are an angle and 
rotating speed of the pendulum, respectively.
Further,
$u_k\in\R$ is an input torque,
$T_s$ is a sampling period, 
and
$\vw_k\in\R^2$
is the external disturbance where
$\vw_k\sim \mathcal{N}(\vmu_w, \mSigma_w)$,
$\vmu_w=
	[\mu_{w,\ang},\mu_{w,\angv}]^\top \in\R^2$ and
$	\mSigma_{w}=\mathrm{diag}(\sigma_{w,\ang}^2,\sigma_{w,\angv}^2)\in\R^{2\times2}$.
Specific values of these and the other variables used in this evaluation are listed in Table~\ref{tb:sim_para} in Appendix.
We let $\vx_k =\left[\ang_k,  \angv_k \right]^{\top}\in\R^2 $ and 
use the following linear approximation model of the above nonlinear system:
\begin{align}
	\vx_{k+1}
	\simeq 
	\left[
	\begin{array}{cc}
		1  & T_{s}\\
		0 & 1 
	\end{array}
	\right]\vx_{k} 
	+
	\left[
	\begin{array}{c}
		0 \\
		T_s\frac{3}{m\ell^2}
	\end{array}
	\right] u_k + \vw_k.\label{eq:sim_nominal}
\end{align}
For simplicity, we let 
\begin{align}
	\vf(\vx)
	=
	\left[
	\begin{array}{c}
		\ang + T_s\angv \\
		\angv -T_s \frac{3g}{2\ell}\sin(\ang+\pi)   
	\end{array}
	\right],\ 
	\mG=
	\left[
	\begin{array}{c}
		0 \\
		T_s\frac{3}{m\ell^2}
	\end{array}
	\right],\ 
	\mA
	=
	\left[
	\begin{array}{cc}
		1  & T_s\\
		0 & 1 
	\end{array}
	\right],\ 
	\mB
	=
	\left[
	\begin{array}{c}
		0 \\
		T_s\frac{3}{m\ell^2}
	\end{array}
	\right]. \nonumber
\end{align}
The approximation errors
$\ve$ in (\ref{approximation error}) is given by
\begin{align}
	\ve(\vx, u) = 
	\vf(\vx) +\mG u - (\mA \vx + \mB u)
	=
	\left[
	\begin{array}{c}
		0 \\
		-T_s \frac{3g}{2\ell}\sin(\ang+\pi)   
	\end{array}
	\right].
\end{align}
In this evaluation,
we set constraints on $\angv_k$ as $-6 \leq \angv_k\leq 6$, $\forall k =1,\ldots,T$.
This condition becomes
\begin{align}
	\vh_1^\top \vx_k \leq d_1,\   
	\vh_2^\top \vx_k \leq d_2,\ \forall k =1,\ldots,T,
\end{align}
where
$\vh_1^\top = [ 0,  1 ]$,   
$\vh_2^\top = [0,   -1]$, 
$d_1=d_2 =6$,   
and $n_c=2$.
Therefore, Assumption~\ref{assumption:3} holds since $\vh_j^\top \mB \neq 0$, $j\in\{1,2\}$.
Furthermore, the approximation model given in (\ref{eq:sim_nominal}) is controllable 
because of its coefficient matrices $\mA$ and $\mB$, 
while its controllability index is $2$.
According to this result,
we set $\tau=2$ and 
we have
\begin{align}
	\sup_{\vx\in \R^2, \ u\in \R} |\vh_j^ \top \ve(\vx, u)| = T_s \frac{3g}{2\ell}, \ \ j\in\{1,2\},\label{eq:sim_bounds}\\
	\sup_{\vx\in \R^2, \ u\in \R} |\vh_j^ \top (\mA+\mI) \ve(\vx, u)| = T_s \frac{3g}{\ell},\ \ 
	j\in\{1,2\},
	\label{eq:sim_bounds_hat}
\end{align}
since $|\sin(\ang+\pi) |\leq 1$,  $\forall \ang \in \R$.
Therefore we used in this evaluation $T_s \frac{3g}{2\ell}$ and $T_s \frac{3g}{2\ell}$
as $\bar{\delta}_j$ and  $\bar{\Delta}_j$, respectively,
and they satisfy Assumption~\ref{assumption:6}.

Regarding immediate cost, we let
\begin{align}
	c_{k+1}= \Bigl( \{(\ang_k +\pi) \bmod 2\pi \} -\pi  \Bigr)^2 +0.1 \angv_k^2 + 0.001 u_k^2.  
\end{align}
The first term corresponds to swinging up the pendulum and keeping it inverted. 
Furthermore, in our method, we used the following conservative inputs:
\begin{align}
	u^{stay}_k =
	- \frac{m\ell^2}{3T_s}(\angv_k + \mu_{w,\ang}),\ 
	\left[
	\begin{array}{c}
		u_k^{back}  \\
		u_{k+1}^{back}
	\end{array}
	\right] &= \left[
	\begin{array}{c}
		- \frac{m\ell^2}{3T_s}(\angv_k +  2 \mu_{w,\ang})  \\
		0
	\end{array}
	\right].
\end{align}
Both of these inputs  satisfy the inequalities in Theorem~\ref{Thm2} with the parameters listed in Table~\ref{tb:sim_para}, 
and thus, they can be used as conservative inputs defined in Definitions~\ref{cond:u_tilde} and \ref{cond:u_back}.

\subsubsection{Four-bar parallel link robot manipulator:}
We let $\vx=[q_1,\ q_2,\ \varpi_1,\ \varpi_2]^\top$ and $\vu=[v_1,\ v_2]^\top$ where 
$q_1$, $q_2$ are angles of links of a robot, $\varpi_1$, $\varpi_2$ are their rotating speed and $v_1$, $v_2$ are armature voltages from an actuator.
The discrete-time dynamics of a robot manipulator with an actuator including external disturbance $\vw_k\in\mathbb{R}^{4}$ 
where $\vw_k\sim \mathcal{N}(\vmu_{w},\mSigma_{w})$, $\vmu_{w} =[\mu_{w,q_1}, \mu_{w,q_2},\mu_{w,{\varpi_1}},\mu_{w,{\varpi_2}}]^\top\in\mathbb{R}^4$ and
$\mSigma_{w} =\mathrm{diag}(\sigma^2_{w,q_1},\sigma^2_{w,q_2},\sigma^2_{w,\varpi_1},\sigma^2_{w,\varpi_2})\in\mathbb{R}^{4\times4}$
is given by
\begin{align}
\vx_{k+1}	
&= 
	\left[\begin{array}{cccc}
		q_{1_k} + T_s \varpi_{1_k}   \\
		q_{2_k} + T_s \varpi_{2_k}   \\
		\varpi_{1_k}   -T_s\frac{\hat{d}_{11}}{\hat{m}_{11}} \varpi_{1_k}
				- T_s\frac{V_{1}}{\hat{m}_{11}} \cos q_{1_k} \\
		\varpi_{2_k}   -T_s\frac{\hat{d}_{22}}{\hat{m}_{22}} \varpi_{2_k} 
				- T_s\frac{V_{2}}{\hat{m}_{22}} \cos q_{2_k} \\
	\end{array}\right]
	+T_s
	\left[\begin{array}{cc}
		0 &0 \\
		0 &0 \\
		\frac{\alpha}{\hat{m}_{11}} & 0 \\
		0 & \frac{\alpha}{\hat{m}_{22}} 
	\end{array}\right]
	\vu_k +\vw_k\cr
&=:
	\vf(\vx_k) +\vg\vu_k +\vw_k,
	\label{eq:mani_SSd}
\end{align}
where
\begin{align*}
&
\hat{m}_{ii} =\eta^2 J_{mi} +M_{ii},\ 
\hat{d}_{ii} = \eta^2 \left(D_{mi}+\frac{K_t K_b}{R}\right),\ i\in\{1,2\},\ 
\alpha= \frac{\eta K_a K_t }{R}.
\end{align*}
The definitions of each symbol in (\ref{eq:mani_SSd}) and their specific values except the sampling period $T_s$ are provided in \cite{namerikawa+95a}. 
Derivation of  (\ref{eq:mani_SSd}) is detailed in Appendix~\ref{append:mani}.  
Similarly, 
we obtain the following linear approximation model of (\ref{eq:mani_SSd}) by ignoring gravity term: 
\begin{align}
	\vx_{k+1}	
	&\simeq 
	\left[\begin{array}{cccc}
	1 & 0 & T_s & 0    \\
	0 & 1 &  0 & T_s    \\
	0 & 0 & (1-T_s\frac{\hat{d}_{11}}{\hat{m}_{11}}) & 0    \\
	0 & 0 & 0  & (1-T_s\frac{\hat{d}_{22}}{\hat{m}_{22}})   \\
	\end{array}\right]
	\left[\begin{array}{c}
	q_{1_k}    \\
	q_{2_k}  \\
	 \varpi_{1_k} \\
	 \varpi_{2_k} \\ 
	\end{array}\right]
+
	T_s
	\left[\begin{array}{cc}
		0 &0 \\
		0 &0 \\
		\frac{\alpha}{\hat{m}_{11}} & 0 \\
		0 & \frac{\alpha}{\hat{m}_{22}} 
	\end{array}\right]
	\vu_k +\vw_k\nonumber\\
	&=:
	\mA\vx_k +\mB\vu_k +\vw_k.
	\label{eq:mani_nom_SSd}
\end{align}
In the same way as the setting of an inverted pendulum problem described above,
we set constraints on the upper and lower bounds regarding rotating speed  $\varpi_1$ and $\varpi_2$ 
with 
$\vh_1 = [0,0,1,0]^{\top}$, $\vh_2 = [0,0,-1,0]^{\top}$, 
$\vh_3 = [0,0,0,1]^{\top}$, $\vh_4 = [0,0,0,-1]^{\top}$. 
Since $|\cos q_{i}|\leq 1$, $i\in\{1,2\}$, 
we have the following relations: 
\begin{align}
	\sup_{\vx\in\mathbb{R}^4,\vu\in\mathbb{R}^2}
	|\vh^\top_j \ve(\vx,\vu)|
	=
	\left\{\begin{array}{cc}
		T_s\frac{V_1}{\hat{m}_{11}},\ & j \in\{1,2\} \\
		T_s\frac{V_2}{\hat{m}_{22}},\ & j \in\{3,4\} 
	\end{array}\right., \\
\sup_{\vx\in\mathbb{R}^4,\vu\in\mathbb{R}^2}
|\vh^\top_j (\mA+\mI) \ve(\vx,\vu)|
=
\left\{\begin{array}{cc}
	|2-T_s\frac{\hat{d}_{11}}{\hat{m}_{11}}|T_s\frac{V_1}{\hat{m}_{11}},\ & j \in\{1,2\} \\
	|2-T_s\frac{\hat{d}_{22}}{\hat{m}_{22}}|T_s\frac{V_2}{\hat{m}_{22}},\ & j \in\{3,4\} 
\end{array}\right.,
\end{align}
We use them as $\bar{\delta}_j$ and $\bar{\Delta}_j$, and therefore Assumption~\ref{assumption:6} holds. 
Assumption~\ref{assumption:3} also holds with $\vh_1, \vh_2, \vh_3, \vh_4$ and $\mB$.
In this setting, we used immediate cost
\begin{align}
	c_{k+1}&= 2\Bigl( \{(q_{1_k} +\pi) \bmod 2\pi \} -\pi  \Bigr)^2 + 2\Bigl( \{((q_{2_k} +\pi)-5\pi/6) \bmod 2\pi \} -\pi  \Bigr)^2  \nonumber \\
	       &\ \ + 0.1 (\varpi_{1_k}^2 +\varpi_{2_k}^2) + 0.001\vu_k^\top \vu_k.  
\end{align}
Furthermore, in our method, we used the following conservative inputs:
\begin{align}
	\vu_{k}^{stay} =
	\left[
\begin{array}{c}
	-\frac{1}{b_1} \{(1-a_1)\varpi_{1_k} + (1-a_1)\mu_{w,\varpi_1}\}  \\
	-\frac{1}{ b_2} \{(1-a_2)\varpi_{2_k} + (1-a_2)\mu_{w,\varpi_2}\}  
\end{array}
\right] ,\\ 
\vu_k^{back}  =
	\left[
	\begin{array}{c}
	-\frac{1}{(1-a_1) b_1} \{(1-a_1)^2\varpi_{1_k} + (2-a_1)\mu_{w,\varpi_1}\}  \\
	-\frac{1}{(1-a_2) b_2} \{(1-a_2)^2\varpi_{2_k} + (2-a_2)\mu_{w,\varpi_2}\}  
	\end{array}
	\right],\ 
\vu_{k+1}^{back}  =
	\left[
	\begin{array}{c}
		0 \\
		0
	\end{array}
	\right],
\end{align}
where $a_1$, $a_2$, $b_1$ and $b_2$ are derived from elements of $\mA$ and $\mB$ and they are given by
$a_1= T_s\hat{d}_{11}/\hat{m}_{11}$,  
$a_2= T_s\hat{d}_{22}/\hat{m}_{22}$,  
$b_1= T_s\alpha/\hat{m}_{11}$, and  
$b_2= T_s\alpha/\hat{m}_{22}$. 
Both of these  inputs  satisfy the inequalities in Theorem~\ref{Thm2} with the parameters listed in Table~\ref{tb:sim_para_Mani}, 
and thus, they can be used as conservative inputs defined in Definitions~\ref{cond:u_tilde} and \ref{cond:u_back}.

\subsubsection{Reinforcement learning algorithm and reference method:}
We have combined our proposed 
safe exploration method (\ref{Proposed_input}) 
with the Deep Deterministic Policy Gradient (DDPG)  algorithm~\cite{Lillicrap+15a} in each experimental setting 
with the immediate costs and conservative inputs described above.
We also combined 
safe exploration method 
given in the previous work~\cite{Okawa} 
that does not take disturbance into account 
with the DDPG algorithm for the reference where 
$u_k^{stay}=0$ as used in that paper.
The network structure and hyperparameters we used throughout this evaluation are listed in Tables~\ref{tb:sim_para} and \ref{tb:sim_para_Mani} in Appendix.

\subsection{Simulation results}\label{subsec:sim_res}

\begin{figure}[t]
	\begin{minipage}{.49\hsize}
		\begin{center}
			\includegraphics[width=.92\textwidth]{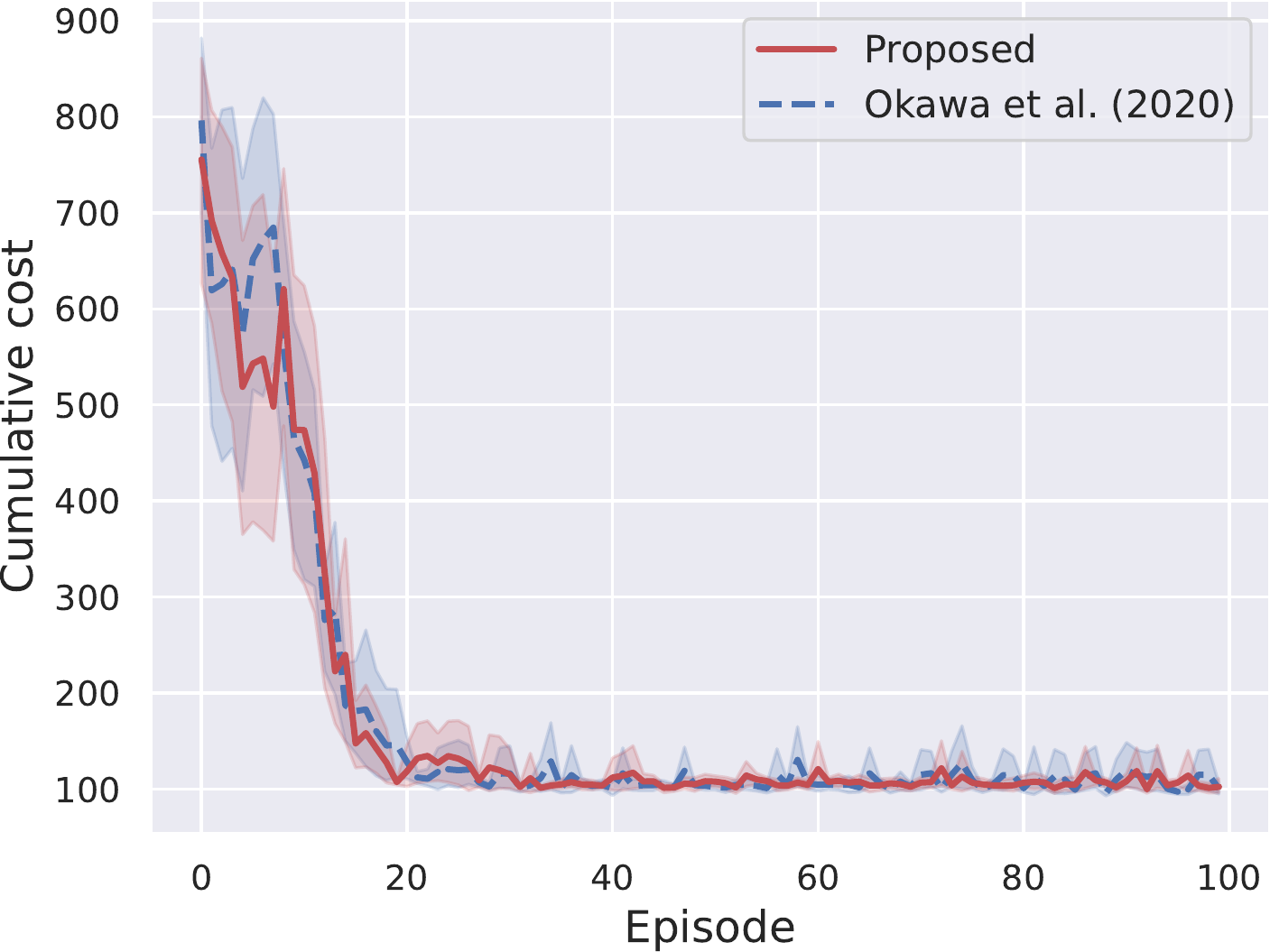}
		\end{center}
	\vspace{2pt}
	\end{minipage}
	\begin{minipage}{.49\hsize}
		\begin{center}
			\includegraphics[width=.98\textwidth]{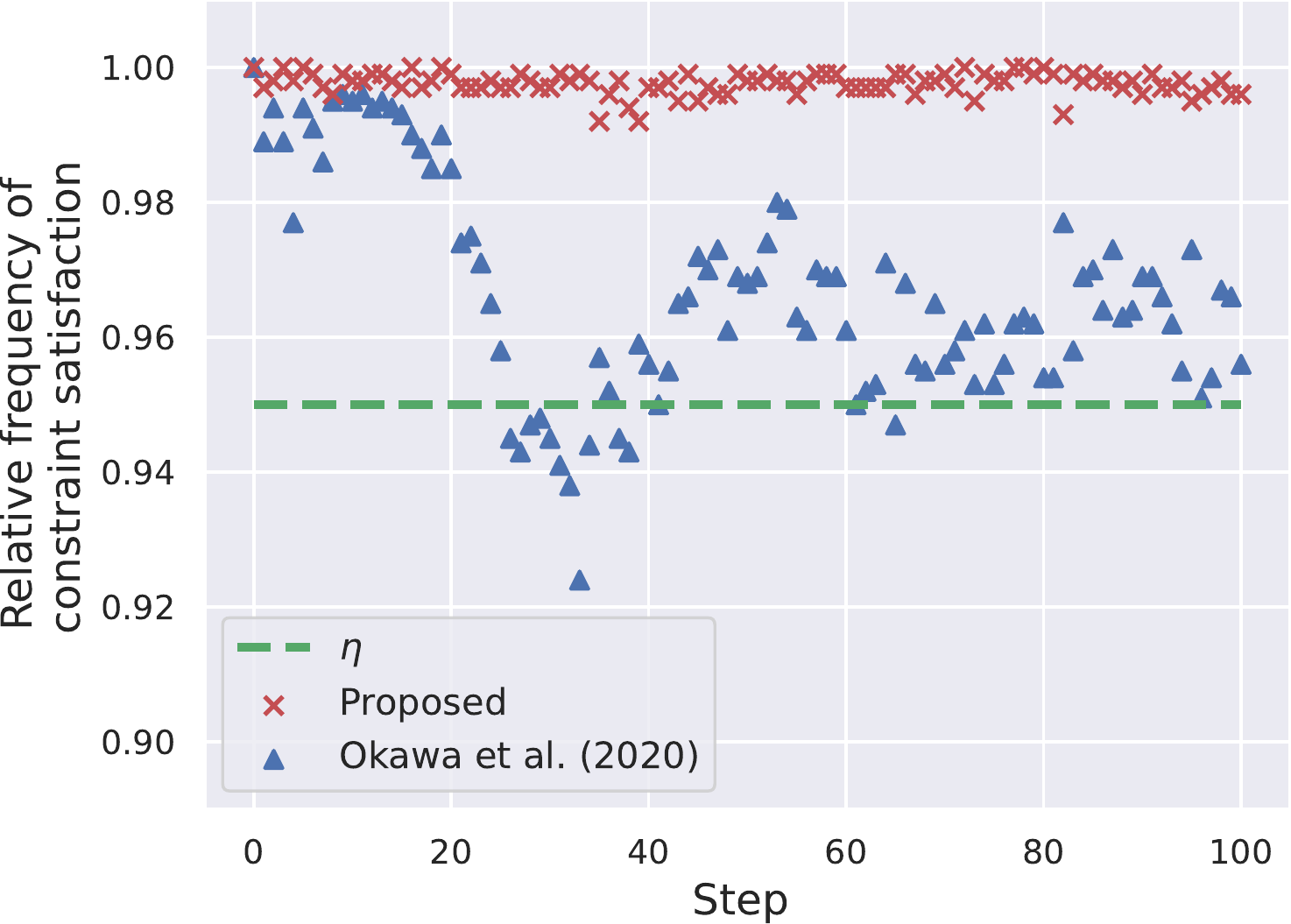}
		\end{center}
	\vspace{2pt}
	\end{minipage}
	\begin{minipage}{.49\hsize}
		\begin{center}
			\includegraphics[width=.92\textwidth]{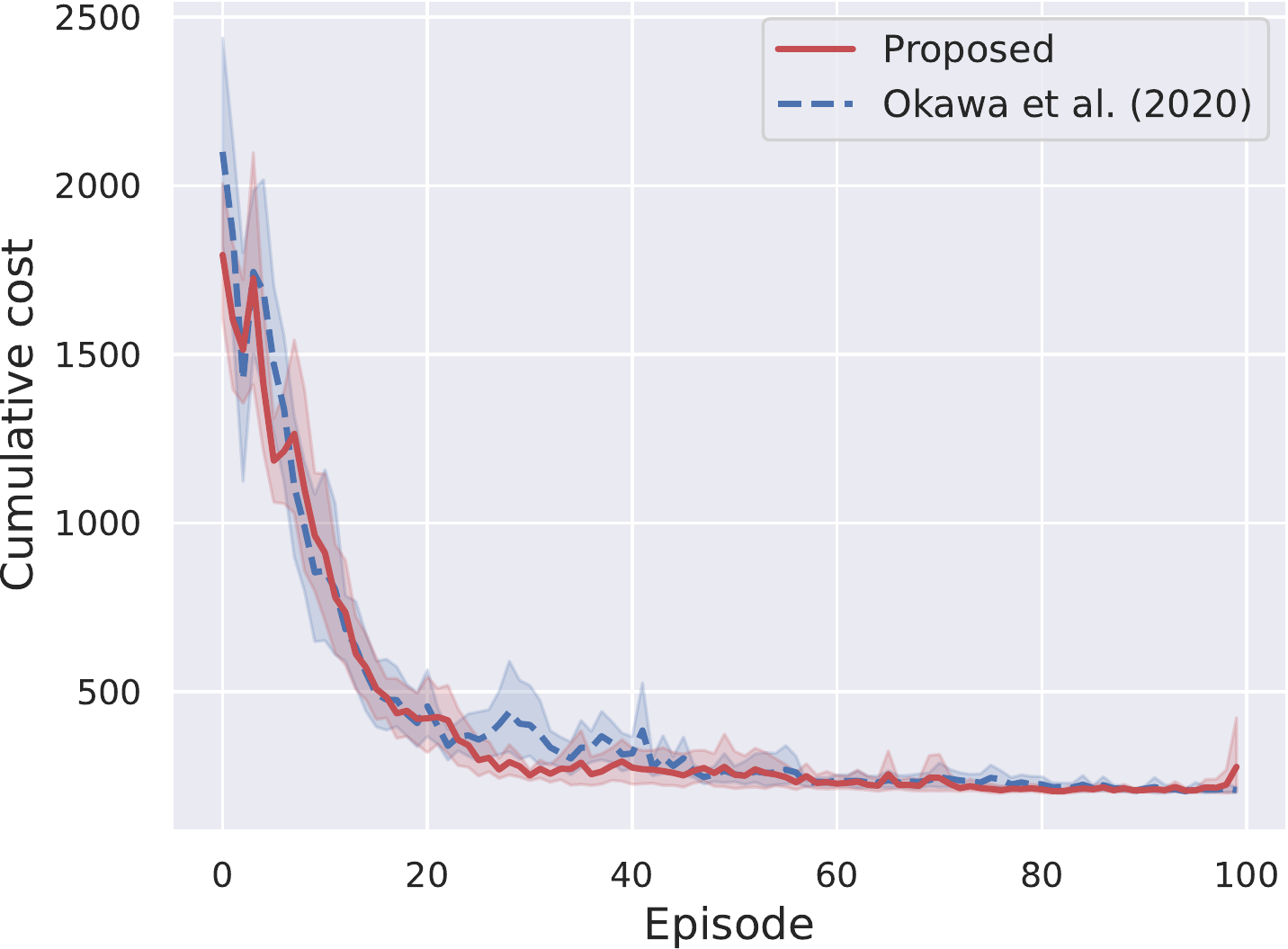}
		\end{center}
	\end{minipage}
	\begin{minipage}{.49\hsize}
		\begin{center}
			\includegraphics[width=.98\textwidth]{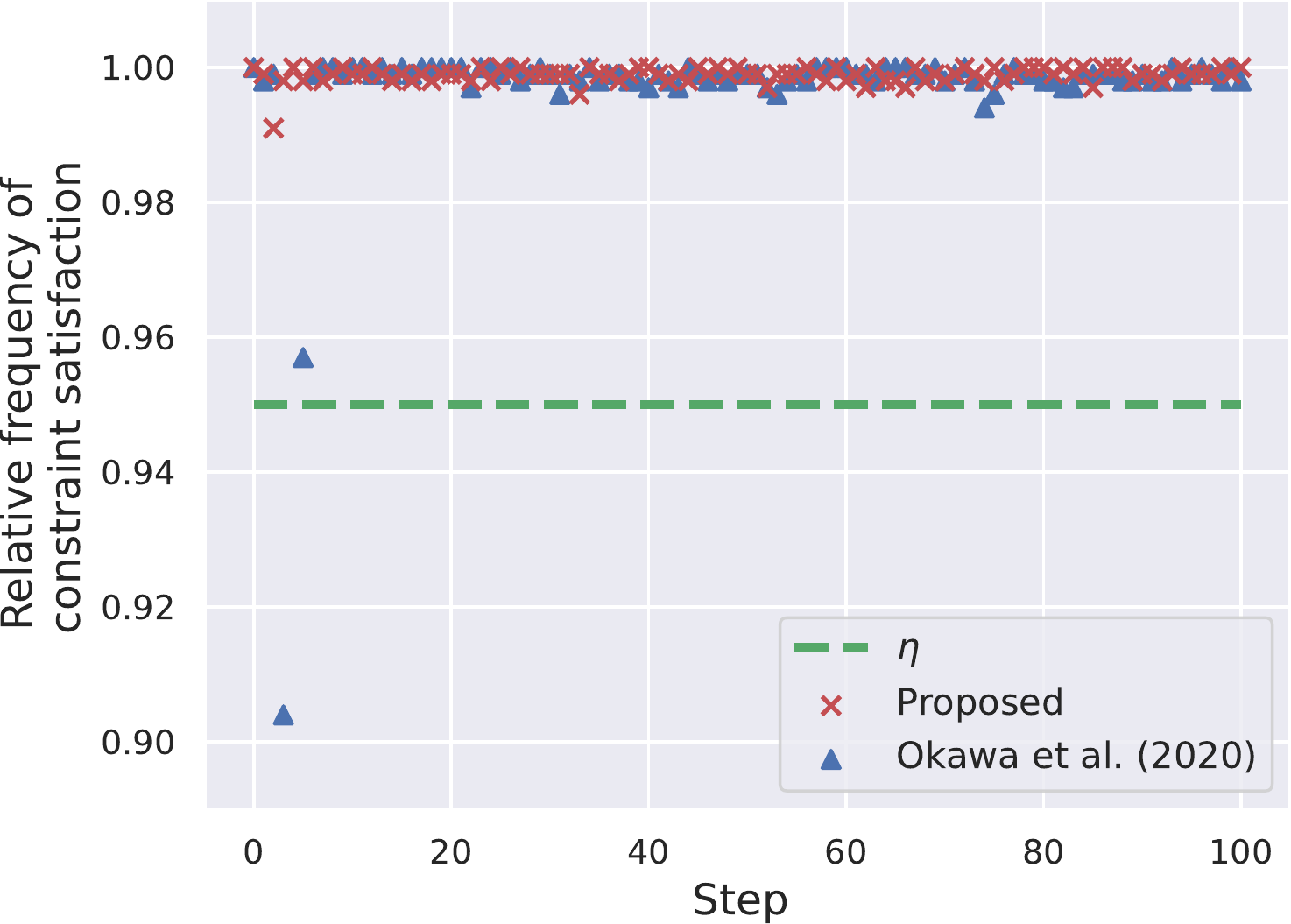}
		\end{center}
	\end{minipage}
	\caption{Results on simulation with ({\bf Top}) an inverted-pendulum and ({\bf Bottom}) a four-bar parallel link robot manipulator: 
		({\bf Left}) Cumulative costs at each episode,
		({\bf Right}) Relative frequencies of constraint satisfaction at each time step. 
		Both the proposed method (red) and the previous one (blue)~\cite{Okawa} enabled to reduce their cumulative costs;
		however, only the proposed method satisfied the relative frequencies of constraint satisfaction to be equal or greater than $\eta$ for all steps in both experimental settings.
	}
	\label{fig:sim_res}
\end{figure}

Figure~\ref{fig:sim_res} shows the results of the cumulative costs at each episode and the relative frequencies of constraint satisfaction.
We evaluated our method and the previous one 
with 100 episodes $\times$  10 runs  of  the simulation (each episode consists of 100 time steps)
under the conditions described in the previous subsection.
We used Intel(R) Xeon(R) CPU E5-2667 v4 @3.20GHz and one NVIDIA V100 GPU. Under this experimental setup, it took about one hour to run through each experiment.
The results shown in both figures are their mean values, while the shaded areas in the top figure show their 95\% confidence intervals.
From these figures, both methods enabled to reduce their cumulative costs as the number of episode increases;
however, only the proposed method satisfied the relative frequencies of constraint satisfaction to be equal or greater than $\eta$ for all steps.

%% file: Sec6.tex

There are two main things we need to care about to use the proposed method.  
First, although it is relaxed compared to the previous work~\cite{Okawa}, the controlled object and disturbance should satisfy several conditions and we need partial prior knowledge about them as described in  Assumptions~\ref{assumption:1} through \ref{assumption:3}. 
In addition, the proposed method requires calculations including  matrices, vectors, nonlinear functions and probabilities. 
This additional computational cost may become a problem if the controller should be implemented as an embedded system.

%% file: Sec7.tex

In this study, 
we proposed a safe exploration method 
for RL to guarantee the safety during learning under the existence of disturbance.
The proposed method
uses partially known information of both the controlled object and disturbances.
We theoretically proved that 
the proposed method achieves the satisfaction of explicit state constraints with a pre-specified probability at every timestep even when the controlled object is exposed to the disturbance following a normal distribution. 
Sufficient conditions to construct conservative inputs used in the proposed method are also provided for its implementation. 
We also experimentally showed the validity and effectiveness of the proposed method through simulation evaluation using an inverted pendulum and a four-bar parallel link robot manipulator.  
%
%
Our future work includes the application of the proposed method to real environments.

%% file: append_proofs.tex
\subsection{Proof of Lemma~\ref{Lem2}}\label{append:proof_Lem2} 
\begin{proof}

By using the definition of $\ve$, that is (\ref{approximation error}), we can rewrite the state equation (\ref{nonlinear system}) as follows:
\begin{align}
\vx_{k+1}=  \mA \vx_k + \mB \vu_k  + \ve(\vx_k, \vu_k) +   \vw_k.
\end{align}

If one $j$ is arbitrarily selected and fixed, the following relation holds for the state $\vx_{k+1}$:
	\begin{align}
	&{\rm Pr}\{\vh_j^{\top}\vx_{k+1}\leq d_j\}\geq q \nonumber \\
	&\Leftrightarrow {\rm Pr}\{\vh_j^{\top}(\mA\vx_k+\mB\vu_k + \ve(\vx_k , \vu_{k}) + \vw_k  )  \leq d_j\}\geq q \nonumber \\
	&\Leftarrow {\rm Pr}\{\vh_j^{\top}(\mA\vx_k+\mB\vu_k+\vw_k)+\delta_j\leq d_j\}\geq q, \ \forall\delta_j\in\{\bar{\delta}_j, \ -\bar{\delta}_j\} \nonumber \\
	&({\rm due \ to \  Assumption}~\ref{assumption:6}) \nonumber \\
	&\Leftrightarrow {\rm Pr}\{\vh_j^{\top}(\mA\vx_k+
	\left[\mB, \mI\right]
	\left[\begin{array}{c}
	\vu_k \\
	\vw_k \\
	\end{array}\right])+\delta_j\leq d_j\}\geq q, \ \forall\delta_j\in\{\bar{\delta}_j, \ -\bar{\delta}_j\} \nonumber \\
	&\Leftrightarrow {\rm Pr}\{\vh_j^{\top}(\mA\vx_k+\mB'\left[\begin{array}{c}
	\vu_k \\
	\vw_k \\
	\end{array}\right])+\delta_j\leq d_j\}\geq q, \ \forall\delta_j\in\{\bar{\delta}_j, \ -\bar{\delta}_j\}. 
	\end{align}
	Input $\vu_k$ and disturbance $\vw_k$ follow normal distributions and are uncorrelated (Assumption~\ref{assumption:2} and (\ref{確率密度関数u})), so if one $\delta_j\in\{\bar{\delta}_j, \ -\bar{\delta}_j\}$ is arbitrarily selected and fixed, the following relation holds \cite{Boyd}:
	\begin{align}
	{\rm Pr}\{\vh_j^{\top}(\mA\vx_k+\mB'\left[\begin{array}{c}
	\vu_k \\
	\vw_k \\
	\end{array}\right])+\delta_j \leq d_j\} \geq q \nonumber \\
	\Leftrightarrow d_j-\vh_j^{\top}\left(\mA\vx_k+\mB'\left[\begin{array}{c}
	\vmu(\vx_k; \vtheta_k) \\
	\vmu_w \\
	\end{array}\right]\right)-\delta_j
	\geq \Phi^{-1}(q)\left\|
	\vh_j^{\top} \mB' \left[\begin{array}{cc}
	\mSigma_{k} &   \\
	& \mSigma_w \\
	\end{array}\right]^\frac{1}{2}
	\right\|_2. 
	\label{eq:cc_relaxation}
	\end{align}
	Hence,
	\begin{align}
	\Phi^{-1}(q)\left\|
	\vh_j^{\top} \mB' \left[\begin{array}{cc}
	\mSigma_{k} &   \\
	& \mSigma_w \\
	\end{array}\right]^\frac{1}{2}
	\right\|_2
	\leq d_j-\vh_j^{\top}(\mA\vx_k+\mB\vmu(\vx_k; \vtheta_k)+\vmu_w)-\delta_j \nonumber \\
	\forall j=1, \ 2, \ \ldots, \ n_c, \ \ \forall\delta_j\in\{\bar{\delta}_j, \ -\bar{\delta}_j\} \nonumber \\ 
	\Rightarrow {\rm Pr}\{\vh_j^{\top}\vx_{k+1}\leq d_j\}\geq q, \ \forall j=1, \ 2, \ \ldots, \ n_c. \label{31}
	\end{align}
	Note that $\Phi^{-1}(q) > 0$ for  $q\in(0.5, \ 1)$. 
	Thus, the inequality on the left side of (\ref{31}) can be rewritten as follows:
	\begin{align}
	\left\|
	\vh_j^{\top} \mB' \left[\begin{array}{cc}
	\mSigma_{k} &   \\
	& \mSigma_w \\
	\end{array}\right]^\frac{1}{2}
	\right\|_2
	\leq \frac{1}{\Phi^{-1}(q)}\left\{d_j-\vh_j^{\top}(\mA\vx_k+\mB\vmu(\vx_k; \vtheta_k)+\vmu_w)-\delta_j\right\}, \nonumber \\
	\forall j=1, \ 2, \ \ldots, \ n_c, \ \ \forall\delta_j\in\{\bar{\delta}_j, \ -\bar{\delta}_j\}. \label{33}
	\end{align}
This completes the proof. \qed
\end{proof}

\subsection{Proof of Lemma~\ref{Lem1}}\label{append:proof_Lem1}  
\begin{proof}
	Consider a discrete-time Markov chain $\{X_k\}$, where $\{1, 2, \ldots, \tau, \tau+1, \tau+2\}$ is the state space and the transition probability matrix with $\rho_i\in(0,1)$, $i=1,2,\ldots,\tau+2$ is as follows: 
\begin{align}
	\left[
	\begin{array}{rrrrrr}
	\rho_1 & 1-\rho_1 & 0 & \cdots & 0 & 0 \\
	\rho_2 & 0 & 1-\rho_2 & \cdots & 0 & 0 \\
	\vdots & \vdots & \vdots & \ddots & \vdots & \vdots \\
	\rho_\tau & 0 & 0 & \cdots & 1-\rho_\tau & 0 \\
	\rho_{\tau+1} & 0 & 0 & \cdots & 0 & 1-\rho_{\tau+1} \\
	\rho_{\tau+2} & 0 & 0 & \cdots & 0 & 1-\rho_{\tau+2}
	\end{array}
	\right].
\end{align}
The state transition diagram is shown in Fig.~\ref{Fig:State_transition}. 
\begin{figure}[hbtp]
	\begin{center}
		\includegraphics[width=.45\textwidth]{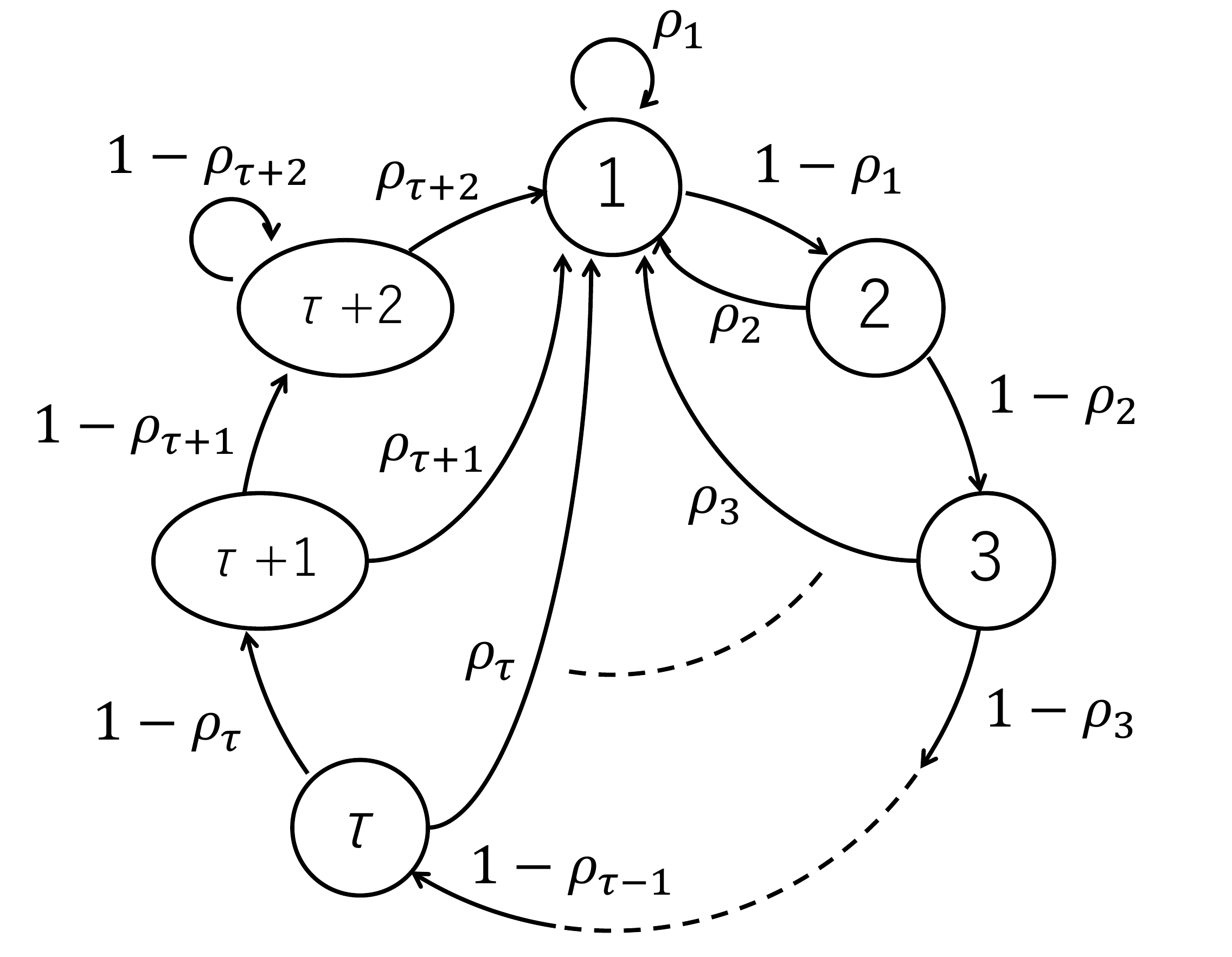}
	\end{center}
	\caption{State transition diagram of the discrete-time Markov chain $\{X_k\}$}
	\label{Fig:State_transition}
\end{figure}\par
\noindent
We use this Markov chain to prove the lemma by relating this to our main problem as follows: 
``$\vx_k\in\gX_s$'' is state $1$, ``$\vx_{k-i}\in\gX_s$ and $\vx_{k-i+1}, \ldots, \vx_k\notin\gX_s$'' is state $i+1$ ($i=1,2,\ldots,\tau$), and ``$\vx_{k -\tau}, \vx_{k-\tau+1}, \ldots, \vx_k\notin\gX_s$'' is state $\tau+2$.
	We define the probability of the state of  Markov chain being $i$ at time $k$ as $p_k^{(i)}$, that is, 
	\begin{align}
	p_k^{(i)}:={\rm Pr}\{X_k=i\}. 
	\end{align}
	We prove by induction that  the inequality 
	\begin{align}
	p_k^{(1)} > \xi^{k}\rho_1^{\tau}
	\end{align}
holds for all $k=1,2, \ldots, T$ when $p_0^{(1)}= {\rm Pr}\{X_0=1\} = {\rm Pr}\{ \vx_0 \in\gX_s \} = 1$. 

	First, consider $k=1, 2, \ldots, \tau$. We have the following relation: 
	\begin{align}
	p_k^{(1)} & \geq p_{0}^{(1)} \rho_1^k  \nonumber \\
	& = \rho_1^k \hspace{4ex} ({\rm because} \ p_0^{(1)}= 1)  \nonumber \\
	&\geq \rho_1^{\tau} \hspace{4ex} ({\rm because} \ k\leq \tau) \nonumber \\
	&> \xi^{k}\rho_1^{\tau} \hspace{4ex} ({\rm because}\ \xi <1 ) \label{eq:induction_1}. 
	\end{align}
	
	Next, consider $k=\tau+1$. From the following two relations in addition to (\ref{eq:induction_1}), 
	\begin{align}
	&p_{\tau+1}^{(1)} = \sum_{i=1}^{\tau+1}\rho_i p_\tau^{(i)}, \\
	&p_\tau^{(i)} = \left(\prod_{j=1}^{i-1}(1-\rho_j)\right)p_{\tau-i+1}^{(1)} \ (i=2, \ldots, \tau+1), 
	\end{align}
	we have the following: 
	\begin{align}
	p_{\tau+1}^{(1)} & = \sum_{i=1}^{\tau+1}\rho_i p_\tau^{(i)} \nonumber \\
	& = \rho_1 p_\tau^{(1)} + \sum_{i=2}^{\tau+1}\rho_i \left(\prod_{j=1}^{i-1}(1-\rho_j)\right)p_{\tau-i+1}^{(1)} \nonumber \\
	& > \rho_1\xi^\tau\rho_1^{\tau}+\sum_{i=2}^{\tau+1}\rho_i\left(\prod_{j=1}^{i-1}(1-\rho_j)\right)\xi^\tau\rho_1^{\tau} \nonumber \\
	&=\left\{ \rho_1+\sum_{i=2}^{\tau+1}\rho_i\prod_{j=1}^{i-1}(1-\rho_j) \right\}\xi^\tau\rho_1^{\tau} \nonumber \\
	&=\!\left\{\! \rho_1+(1-\rho_1)\!\left(\!\rho_2+\sum_{i=3}^{\tau+1}\rho_i\prod_{j=2}^{i-1}(1-\rho_j)\!\right)\!\right\}\!\xi^\tau\rho_1^{\tau}. \label{Lem1HalfWay}
	\end{align}
	From Definition~\ref{cond:u_back}, the probability of moving inside $\gX_s$ within $\tau$ steps is greater than or equal to $\xi$ if we use $\vu_{k}^{back}, \ \vu_{k+1}^{back}, \ \ldots, \vu_{k+\tau-1}^{back}$ when $\vx_{k-1} \in \gX_s$ and $\vx_{k} \notin \gX_s$ occur for $k\geq 1$. 
This is rewritten in the following form: 
	\begin{align}
	\rho_2+\sum_{i=3}^{\tau+1}\rho_i\prod_{j=2}^{i-1}(1-\rho_j)\geq\xi. \label{Lem1HalfWay2}
	\end{align}
From (\ref{Lem1HalfWay}) and (\ref{Lem1HalfWay2}), we have
\begin{align}
	p_{\tau+1}^{(1)} &\geq\left\{\rho_1+(1-\rho_1)\xi\right\}\xi^\tau\rho_1^{\tau} \nonumber \\
	&=\left\{ \xi + \rho_1(1-\xi) \right\}\xi^\tau\rho_1^{\tau} \nonumber \\
	&> \xi^{\tau+1}\rho_1^{\tau}.
\end{align}
	Therefore, $p_k^{(1)} > \xi^{k}\rho_1^{\tau}$ holds also at $k=\tau+1$. 
	
	Suppose that $p_k^{(1)} > \xi^{k}\rho_1^{\tau}$ holds for $k\geq \tau+1$. The following recurrence formulas hold:
	\begin{align}
	\left\{
	\begin{aligned}
	&p_{k+1}^{(1)}=\sum_{i=1}^{\tau+2}\rho_ip_k^{(i)}, \\
	&p_{k+1}^{(i)}=(1-\rho_{i-1})p_k^{(i-1)} \ \ (i=2, 3, \ldots, \tau+1), \\
	&p_{k+1}^{(\tau+2)}=(1-\rho_{\tau+1})p_k^{(\tau+1)}+(1-\rho_{\tau+2})p_k^{(\tau+2)}. 
	\end{aligned}
	\right.
	\end{align}
	From
	\begin{align}
	p_k^{(i)}=\left(\prod_{j=1}^{i-1}(1-\rho_j)\right)p_{k-i+1}^{(1)},
	\end{align}
	we obtain the following relation:
	\begin{align}
	p_{k+1}^{(1)}&= \sum_{i=1}^{\tau+2}\rho_ip_k^{(i)} \nonumber \\
	&> \sum_{i=1}^{\tau+1}\rho_ip_k^{(i)} \nonumber \\
	&=
	\rho_1p_k^{(1)}+\sum_{i=2}^{\tau+1}\rho_i\left(\prod_{j=1}^{i-1}(1-\rho_j)\right)p_{k-i+1}^{(1)} \nonumber \\
	&> \rho_1\xi^k\rho_1^{\tau+1}+\sum_{i=2}^{\tau+1}\rho_i\left(\prod_{j=1}^{i-1}(1-\rho_j)\right)\xi^{k}\rho_1^{\tau} \nonumber \\
	&=\left\{ \rho_1+\sum_{i=2}^{\tau+1}\rho_i\prod_{j=1}^{i-1}(1-\rho_j) \right\}\xi^{k}\rho_1^{\tau} \nonumber \\
	&\geq\left\{\rho_1+(1-\rho_1)\xi\right\}\xi^{k}\rho_1^{\tau} \nonumber \\
	&=\left\{ \xi + \rho_1(1-\xi) \right\}\xi^{k}\rho_1^{\tau} \nonumber \\
	&> \xi^{k+1}\rho_1^{\tau}.
	\end{align}
	Hence, $p_k^{(1)} > \xi^{k}\rho_1^{\tau}$ holds also at $k+1$.
	
	Therefore, $p_k^{(1)}>\xi^{k}\rho_1^{\tau}$ hold for all $k=1,2, \ldots, T$.
	Note that $\rho_{1}=p$ because of the definitions of $\rho_1$ and $p$.
	This concludes that $p_k^{(1)} > \xi^{k}p^{\tau}$ holds for all $k=1,2,\ldots,T$, and thus
	${\rm Pr}\{\vx_{k}\in\gX_s\} \geq \xi^{k}p^{\tau}$ also holds for all $k=1,2,\ldots,T$.
	Now the lemma is proved. \qed
\end{proof}

\subsection{Proof of Theorem~\ref{Thm2}}\label{append:proof_Thm2}
\begin{proof}
	First, from Bonferroni's inequality, the following relation holds for $q'=1-\frac{1-q}{n_c}$:
	\begin{align}
		&{\rm Pr}\{\mH\vx_{k+1}\preceq \vd\} \geq q \Leftarrow {\rm Pr}\{\vh_j^{\top}\vx_{k+1}\leq d_j\}\geq q', \ \ 
		\forall j=1, \ldots, n_c. \label{theorem3-1}
	\end{align}
	Hence,
	\begin{align}
		&{\rm Pr}\{\mH\vx_{k+1}\preceq \vd\} \geq q \nonumber \\
		&\Leftarrow {\rm Pr} \left\{ \vh_j^{\top} \left(\mA\vx_k+\mB\vu_k+\ve_k+\vw_k\right) \leq d_j \right\}\geq q', \ \forall j=1, \ldots, n_c \nonumber \\
		&\Leftarrow {\rm Pr} \left\{ \vh_j^{\top} \left(\mA\vx_k+\mB\vu_k+\vw_k\right)+\delta_j  \leq d_j \right\}\geq q', \ \forall \delta_j, \ \forall j=1, \ldots, n_c, 
		\label{theorem3-2}
	\end{align}
	where $\ve_k := \ve(\vx_k, \vu_k)$. 
	Next, as in the proof of Lemma~\ref{Lem2}, the following relation holds: 
	\begin{align}
		&{\rm Pr} \{ \vh_j^{\top} \left(\mA\vx_k+\mB\vu_k+\vw_k\right)+\delta_j  \leq d_j \}\geq q' \nonumber \\
		&\Leftrightarrow d_j - \vh_j^{\top}\left(\mA\vx_k+\mB\vu_k\right) -\delta_j - \vh_j^{\top}\vmu_w \geq 
		\Phi^{-1}(q')\left\| \vh_j^{\top} \mSigma^{\frac{1}{2}}_w\right\|_2. \label{theorem3-3}
	\end{align}
	Therefore, the first part of the theorem is proved.
	
	The state $\vx_{k+\tau}$ can be expressed by $\vx_{k+\tau-i}$, $i=1, 2, \ldots, \tau$ as follows: 
	\begin{align}
		\vx_{k+\tau}
		=&\mA\vx_{k+\tau-1}+\mB\vu_{k+\tau-1}+\ve_{k+\tau-1}+\vw_{k+\tau-1} \nonumber \\
		=&\mA\left(\mA\vx_{k+\tau-2}+\mB\vu_{k+\tau-2}+\ve_{k+\tau-2}+\vw_{k+\tau-2} \right) \nonumber \\
		 &+\mB\vu_{k+\tau-1}+\ve_{k+\tau-1}+\vw_{k+\tau-1} \nonumber \\
		=&\mA^2\vx_{k+\tau-2}+\left[\mA\mB,  \mB\right] \left[\vu_{k+\tau-2}^{\top}, \vu_{k+\tau-1}^{\top}\right]^{\top} \cr
		 &+\left[\mA, \mI\right] \left[\ve_{k+\tau-2}^{\top},  \ve_{k+\tau-1}^{\top} \right]^{\top} +\left[\mA,  \mI\right] \left[\vw_{k+\tau-2}^{\top} \  \vw_{k+\tau-1}^{\top} \right]^{\top} \nonumber \\
		\vdots& \nonumber \\
		=&\mA^\tau {\vx_k} + \hat{\mB} \mU_k + \hat{\mC} \mE_k + \hat{\mC} \mW_k, \label{x{k+tau}}
	\end{align}
	where
	\begin{align*}
		&\hat{\mB} := [\mA^{\tau-1}\mB, \mA^{\tau-2}\mB,  \ldots, \mB],\    
		\hat{\mC} := [\mA^{\tau-1}, \mA^{\tau-2},  \ldots, \mI],   \\
		&\mU_k := [\vu_k^{\top}, \vu_{k+1}^{\top}, \ldots, \vu_{k+\tau-1}^{\top}]^{\top},\    
		\mE_k := [\ve_k^{\top}, \ve_{k+1}^{\top},  \ldots,  \ve_{k+\tau-1}^{\top}]^{\top},\\  
		&\mW_k := [\vw_k^{\top},  \vw_{k+1}^{\top},  \ldots,  \vw_{k+\tau-1}^{\top}]^{\top}.  
	\end{align*}
	From Bonferroni's inequality, we have
	\begin{align}
		&{\rm Pr}\{\mH\vx_{k+\tau}\preceq \vd\} \geq q \Leftarrow {\rm Pr}\{\vh_j^{\top}\vx_{k+\tau}\leq d_j\}\geq q', \ \ 
		\forall j=1, \ldots, n_c. \label{theorem2-1}
	\end{align}
	Next, as in the proof of Lemma~\ref{Lem2}, the following relation holds:
	\begin{align}
		&{\rm Pr} \{ \vh_j^{\top} \left(\mA^\tau \vx_k + \hat{\mB} \mU_k + \hat{\mC} \mW_k\right)+\Delta_j \leq d_j \}\geq q' \nonumber \\
		&\Leftrightarrow d_j \!-\! \vh_j^{\top}\left(\mA^\tau \vx_k \!+\! \hat{\mB} \mU_k \right) \!-\!\Delta_j \!-\! \vh_j^{\top} \hat{\mC} \hat{\vmu}_w \!\geq\! 
		\Phi^{-1}(q')\!\left\| \vh_j^{\top} \hat{\mC} \!\left[
		\begin{array}{ccc}
			\mSigma_w & &  \\
			& \ddots & \\
			& & \mSigma_w
		\end{array}
		\right]^\frac{1}{2}\right\|_2, \label{theorem2-3}
	\end{align}
	where $\hat{\vmu}_w=\left[\vmu_w^{\top}, \ldots, \vmu_w^{\top} \right]^{\top} \in \mathbb{R}^{n \times \tau}$.
	Therefore, the second part of the theorem is proved. \qed
\end{proof}

\subsection{Proof of Theorem~\ref{Thm1}}\label{append:proof_Thm1} 
\begin{proof}
	First, consider the case of $(\mathrm{i})$ in (\ref{Proposed_input}). 
Remember that $\xi$ is a positive real number such that $\eta^{\frac{1}{T}}<\xi < 1$. 
From this inequality, we have $\eta<\xi^{T}  $. We also have $\xi^{T}  \le \xi^{k}$ for all $k = 1,2, \dots, T $. 
Thus we have
\begin{align}
\frac{\eta}{\xi^{k}} < 1. 
\end{align}
The parameter $\eta $ is selected from the interval $(0.5, 1)$. We also have $\xi^{k} < 1$. Thus we have
\begin{align}
0.5 < \eta < \frac{\eta}{\xi^{k}}.  \label{eq:eta_1}
\end{align}
Therefore, since $\tau$ is a positive integer, the following relationship holds:
\begin{align}
0.5 < \frac{\eta}{\xi^{k}} \leq \left( \frac{\eta}{\xi^{k}} \right)^{\frac{1}{\tau}} < 1. \label{eq:eta_2}
\end{align}
This leads to the following:
\begin{align}
0.5 < \eta_{k}^{\prime} = 1 - \frac{1- \left(\frac{\eta}{\xi^k} \right)^{\frac{1}{\tau}}}{n_c} < 1. \label{eq:eta_prime}
\end{align}

Note that $\vh_j^{\top} \mB \neq \boldsymbol{0}, \forall j=1,2,\ldots,n_c$ (Assumption~\ref{assumption:3}), and the left side of (\ref{Proposed_Sigma}) is rewritten as follows:
\begin{align}
	\left\|
	\vh_j^{\top} \mB' \left[\begin{array}{cc}
	\mSigma_{k} &   \\
	& \mSigma_{w} 
	\end{array}\right]^\frac{1}{2}
	\right\|_2
= \left \| \left[\vh_j^{\top} \mB \mSigma_{k}^{\frac{1}{2}}  ,\ \vh_j^{\top} \mSigma_{w}^{\frac{1}{2}}  \right] \right \|_{2.}
\end{align}
Thus, when 
\begin{align}
\left\|\vh_j^{\top} \mSigma_w^\frac{1}{2}
	\right\|_2 \leq \frac{1}{\Phi^{-1}(\eta_{k}^{\prime})}(d_j-\vh_j^{\top}\hat{\vx}_{k+1}-\delta_j),  \forall \delta_j \in \{\pm \bar{\delta}_j\},  \forall j=1,\ldots,n_c
\end{align}
holds, 
there exists feasible solutions for  (\ref{Proposed_Sigma}). 

Therefore, from Lemma~\ref{Lem2}, if the input is determined by 
(\ref{確率密度関数u}) with $\mSigma_{k}$ satisfying  (\ref{Proposed_Sigma}), 
the following inequality holds:
		\begin{align}
		{\rm Pr}\{\vh_j^{\top}\vx_{k+1}\leq d_j\}\geq \eta_{k}^{\prime}, \ \ \forall j=1, \ldots, n_c. \label{38}
		\end{align}
	Note that (\ref{38}) means that the probability that each state constraint is satisfied is larger than or equal to $\eta_{k}^{\prime}$, while (\ref{chance constraint}) means that the probability that all state constraints are satisfied at the same time is greater than or equal to a certain value.
	Here, from Bonferroni's inequality we have
	\begin{align}
	{\rm Pr}\{\vh_j^{\top}\vx_{k+1} \leq d_j, \ \  \forall j=1, \ldots, n_c\} \geq \sum_{j=1}^{n_c}{\rm Pr}\{\vh_j^{\top}\vx_{k+1} \leq d_j\}-(n_c-1).
	\end{align}
	Therefore, from (\ref{38}) and the definition of $\eta_k^{\prime}$ written in (\ref{eq:eta_prime}),
	\begin{align}
	&{\rm Pr}\{\vh_j^{\top}\vx_{k+1} \leq d_j\}\geq 1-\frac{1-\left(\frac{\eta}{\xi^{k}}\right)^{\frac{1}{\tau}}}{n_c}, \ \ \forall j=1, \ldots, n_c \nonumber \\
	&\Rightarrow \sum_{j=1}^{n_c}{\rm Pr}\{\vh_j^{\top}\vx_{k + 1}\leq d_j\}\geq n_c-(1-\left(\frac{\eta}{\xi^{k}}\right)^{\frac{1}{\tau}}) \nonumber \\
	&\Leftrightarrow \sum_{j=1}^{n_c}{\rm Pr}\{\vh_j^{\top}\vx_{k + 1}\leq d_j\}-(n_c-1) \geq \left(\frac{\eta}{\xi^{k}}\right)^{\frac{1}{\tau}} \nonumber \\
	&\Rightarrow {\rm Pr}\{\vh_j^{\top}\vx_{k+1}\leq d_j, \ \  \forall j=1, \ldots, n_c\}\geq \left(\frac{\eta}{\xi^{k}}\right)^{\frac{1}{\tau}} \nonumber \\
	&\Leftrightarrow {\rm Pr}\{\mH\vx_{k+1}\preceq \vd\} \geq \left(\frac{\eta}{\xi^{k}}\right)^{\frac{1}{\tau}} \label{cc'bon}
	\end{align}
	holds. That is, (\ref{38}) is a sufficient condition for
	\begin{align}
	{\rm Pr}\{\mH\vx_{k+1}\preceq \vd\}\geq\left(\frac{\eta}{\xi^{k}}\right)^{\frac{1}{\tau}}. \label{機会制約１_append}
	\end{align}
	Hence, when we determine input by 
	(\ref{確率密度関数u}) with $\mSigma_{k}$ satisfyin  (\ref{Proposed_Sigma}), 
	chance constraints (\ref{chance constraint}) are satisfied for $k = 1,2, \dots, T$.
	
	Next, in the case of $(\mathrm{ii})$ in (\ref{Proposed_input}), by determining input as $\vu_k=\vu_{k}^{stay}$, ${\rm Pr}\{\mH\vx_{k+1}\preceq \vd\}\geq\left(\frac{\eta}{\xi^{k}}\right)^{\frac{1}{\tau}}$ holds from Definition~\ref{cond:u_tilde}.
	Therefore, when $\vx_k\in\gX_s$, ${\rm Pr}\{\mH\vx_{k+1}\preceq \vd\}\geq\left(\frac{\eta}{\xi^{k}}\right)^{\frac{1}{\tau}}$ holds by determining input $\vu_k$ according to (\ref{Proposed_input}).

	Finally, by determining input as $\vu_k=\vu_k^{back}$ in case $(\mathrm{iii})$ of (\ref{Proposed_input}), ${\rm Pr}\{\mH\vx_k\preceq \vd\}\geq \eta$ holds for any $\vx_k\in\R^n$, $k=1,2,\ldots,T$ due to Lemma~\ref{Lem1}. 
	Hence, noting $\left(\frac{\eta}{\xi^{k}}\right)^{\frac{1}{\tau}} > \eta$ as shown in (\ref{eq:eta_1}) and (\ref{eq:eta_2}), ${\rm Pr}\{\mH\vx_k\preceq \vd\}\geq \eta$ is satisfied for all time $k=1,2, \ldots, T$. \qed
\end{proof}


%% file: append_sim.tex
\subsection{Configuration figures of experimental setup}\label{append:fig}
\begin{figure}[htbp]
	\begin{minipage}{.49\hsize}
		\begin{center}
			\includegraphics[width=.6\textwidth]{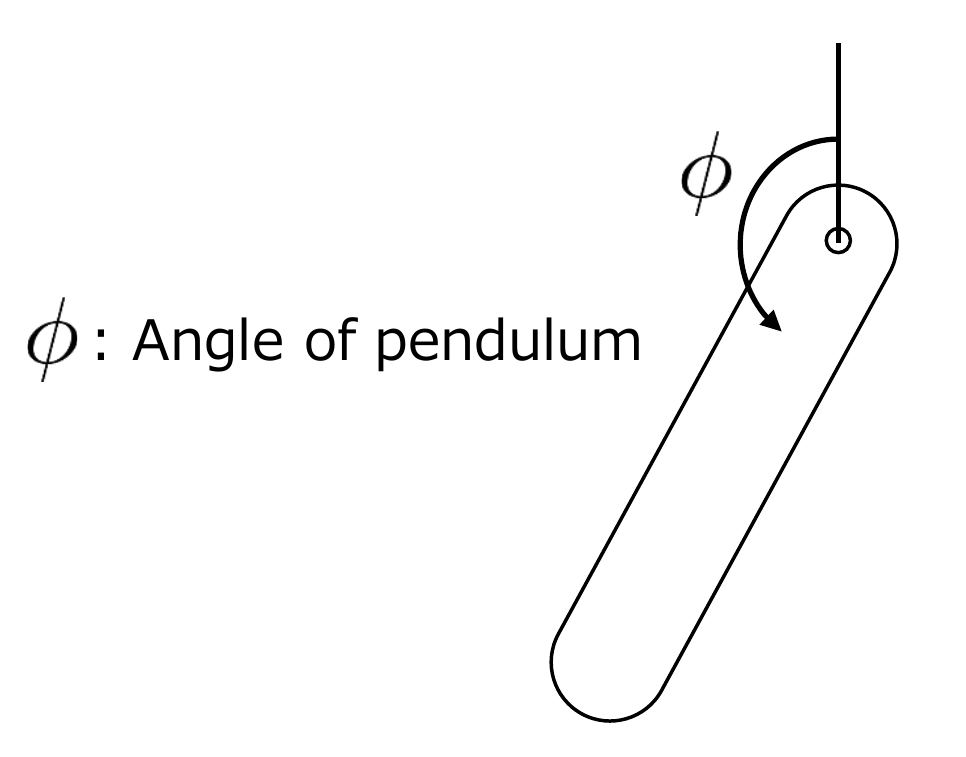}
		\end{center}
	\end{minipage}
	\begin{minipage}{.4\hsize}
		\begin{center}
			\includegraphics[width=1.2\textwidth]{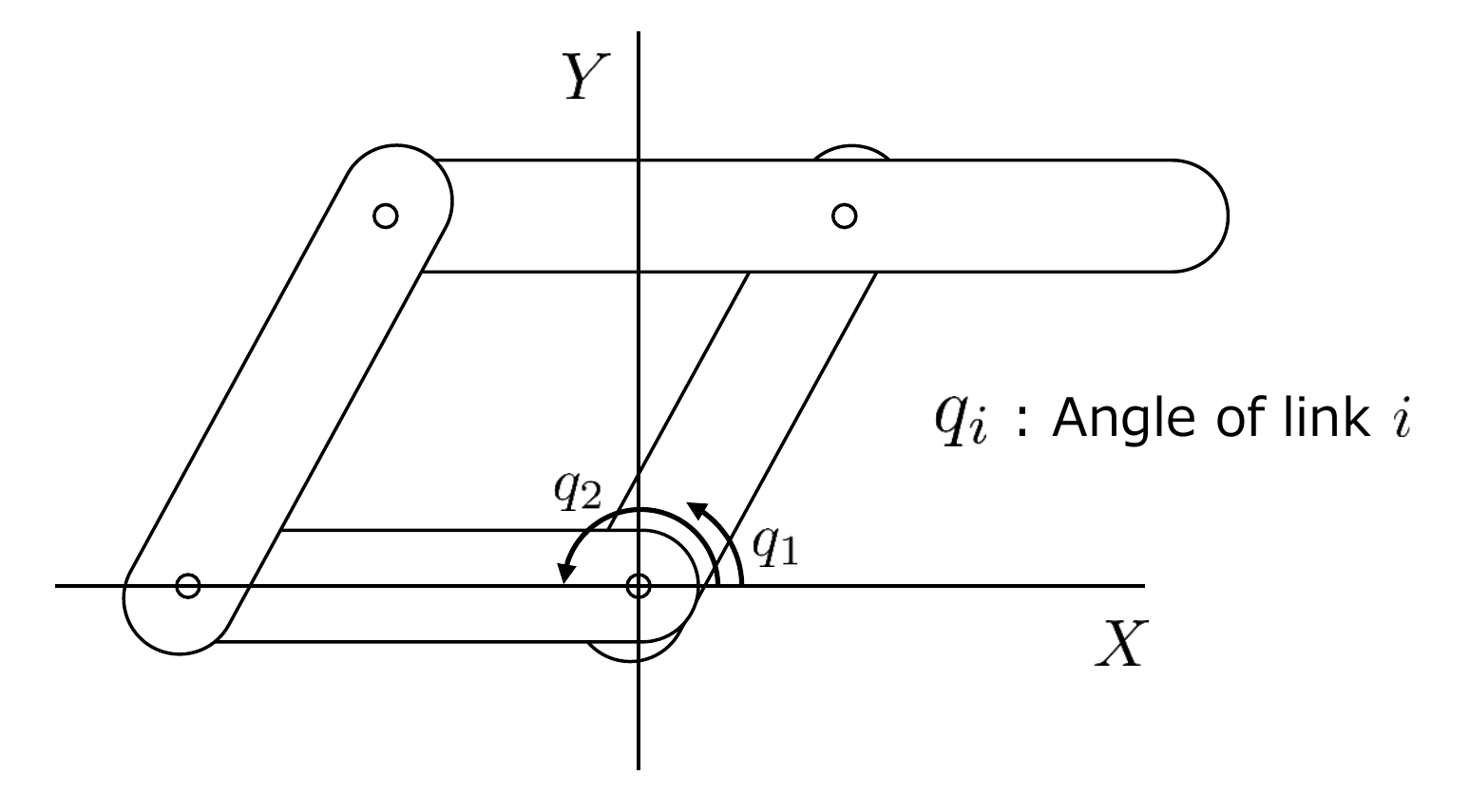}
		\end{center}
	\end{minipage}
	\caption{Configuration figures of ({\bf Left}) an inverted-pendulum  and ({\bf Right}) a four-bar parallel link robot manipulator 
	}
	\label{fig:sim_fig}
\end{figure}

Simplified configuration figures of an inverted pendulum and a four-bar parallel link robot manipulator we used in our verification in Section~\ref{sec:sim_pend} are displayed in Fig.~\ref{fig:sim_fig}. 
We refer the readers to \cite{namerikawa+95a} for the detailed figures of the robot manipulator.

\subsection{Details of the dynamics of robot manipulator}\label{append:mani}
In this appendix, we describe how to derive the discrete-time state space equation (dynamics) of a four-bar parallel link robot manipulator with an actuator given in (\ref{eq:mani_SSd}) from its continuous-time dynamics.

According to Namerikawa et al.~\cite{namerikawa+95a},  the continuous-time dynamics of this robot manipulator with an actuator is given by
\begin{align}
\left[\begin{array}{cc}
	\hat{m}_{11} & \hat{m}_{12} \\
	\hat{m}_{21} & \hat{m}_{22} \\
\end{array}\right]
\left[\begin{array}{c}
	\ddot{q}_{1}  \\
	\ddot{q}_{2}  \\
\end{array}\right]
+
\left[\begin{array}{cc}
	\hat{d}_{11} & \hat{d}_{12} \\
	\hat{d}_{21} & \hat{d}_{22} \\
\end{array}\right]
\left[\begin{array}{c}
	\dot{q}_{1}  \\
	\dot{q}_{2}  \\
\end{array}\right]
+
\left[\begin{array}{c}
	\hat{g}_{1}  \\
	\hat{g}_{2}  \\
\end{array}\right]
=
\alpha
\left[\begin{array}{c}
	v_{1}  \\
	v_{2}  \\
\end{array}\right],
\label{eq_app:mani}
\end{align}
where
\begin{align*}
&
\hat{m}_{ii} =\eta^2 J_{mi} +M_{ii},\ i\in\{1,2\},\ 
\hat{m}_{12}=\hat{m}_{21}=M_{12} \cos (q_2-q_1),\ \\
&
\hat{d}_{ii} = \eta^2 \left(D_{mi}+\frac{K_t K_b}{R}\right),\ i\in\{1,2\},\\
&
\hat{d}_{12} = -M_{12} \dot{q}_2\sin(q_2-q_1),\ 
\hat{d}_{21} = M_{12} \dot{q}_1\sin(q_2-q_1),\ \\
&
\hat{g}_1=V_1 \cos q_1,\ \hat{g}_2 =V_2 \cos q_2,\ 
\alpha= \frac{\eta K_a K_t }{R}.
\end{align*}
From the definitions of each symbol in the above equation and their values provided in~\cite{namerikawa+95a},
$\hat{m}_{11}$, $\hat{m}_{22}$,  $\hat{d}_{11}$, $\hat{d}_{22}$, and $\alpha$ are constant model parameters.
In addition, it is obvious that $\left|\hat{m}_{11} \right|$ and $\left|\hat{m}_{22} \right|$ are far larger than $\max\left| \hat{m}_{12}(q) \right|$ and $\max\left| \hat{m}_{21}(q) \right|$, 
and 
$| \hat{d}_{11} |$, $| \hat{d}_{22} |$  are larger than $\max|\hat{d}_{12}(q,\dot{q})|$ and $\max|\hat{d}_{12}(q,\dot{q})|$, if $\dot{q}_{i} \simeq 1,\ i\in\{1,2\}$.
Therefore, by ignoring these interference terms, we have
\begin{align}
\left[\begin{array}{cc}
	\hat{m}_{11} & 0 \\
	0 & \hat{m}_{22} \\
\end{array}\right]
\left[\begin{array}{c}
	\ddot{q}_{1}  \\
	\ddot{q}_{2}  \\
\end{array}\right]
+
\left[\begin{array}{cc}
	\hat{d}_{11} & 0 \\
	0 & \hat{d}_{22} \\
\end{array}\right]
\left[\begin{array}{c}
	\dot{q}_{1}  \\
	\dot{q}_{2}  \\
\end{array}\right]
+
\left[\begin{array}{c}
	\hat{g}_{1}  \\
	\hat{g}_{2}  \\
\end{array}\right]
=
\alpha
\left[\begin{array}{c}
	v_{1}  \\
	v_{2}  \\
\end{array}\right].
\label{eq_app:mani2}
\end{align}
Consequently, the continuous-time state equation of the robot manipulator with an actuator becomes
\begin{align}
	\frac{d}{dt}
	\left[\begin{array}{c}
		{q}_{1}  \\
		{q}_{2}  \\
		\dot{q}_{1}  \\
		\dot{q}_{2}  \\
	\end{array}\right]
=
	\left[\begin{array}{cccc}
		\dot{q}_{1}  \\
		\dot{q}_{2}  \\
		-\frac{\hat{d}_{11}}{\hat{m}_{11}} \dot{q}_1 - \frac{V_{1}}{\hat{m}_{11}} \cos q_1 \\
		-\frac{\hat{d}_{22}}{\hat{m}_{22}} \dot{q}_2 - \frac{V_{2}}{\hat{m}_{22}} \cos q_2 \\
	\end{array}\right]
	+
	\left[\begin{array}{cc}
		0 &0 \\
		0 &0 \\
		\frac{\alpha}{\hat{m}_{11}} & 0 \\
		0 & \frac{\alpha}{\hat{m}_{22}} 
	\end{array}\right]
	\left[\begin{array}{c}
		v_{1}  \\
		v_{2}  \\
	\end{array}\right].
	\label{eq_app:mani_SSc}
\end{align}
Now, we redefine  $\dot{q}_{i}$ as $\varpi_i$, $i\in\{1,2\}$, and let $\vx=[q_1,\ q_2,\ \varpi_1,\ \varpi_2]^\top$ and $\vu=[v_1,\ v_2]^\top$.
By applying the Euler method to (\ref{eq_app:mani_SSc}) with a sampling period $T_s$, we obtain the discrete-time dynamics of the robot manipulator given in (\ref{eq:mani_SSd}).

\subsection{Simulation parameters and hyperparameteres}\label{append:sim}
Simulation parameters we used in our verification with an inverted pendulum and a four-bar parallel link robot manipulator are listed in Tables~\ref{tb:sim_para} and \ref{tb:sim_para_Mani}, respectively.
In both experiment, we used hyperparameters lised in Table~\ref{tb:sim_para_ddpg} for the DDPG algorithm with Adam to train networks.
In addition, we used the same network structure given in code examples from Keras\footnote[1]{The code is available \url{https://github.com/keras-team/keras-io/blob/master/examples/rl/ddpg_pendulum.py} with th Apache License 2.0}. 
\input{tb_sim_paras}

%% file: tb_sim_paras.tex
\begin{table}[htbp]
	\caption{Simulation parameters (Inverted pendulum)}
	\label{tb:sim_para}
	\centering
		\begin{tabular}{|l|l|l|}
        \hline
		Symbol  &Definition &Value \\
        \hline
			$T$ & Number of simulation steps & $100$	\\
			$N$ & Number of learning episodes & $100$	\\
			$m$ & Mass (kg) & $1$ \\
			$\ell$ & Length of pendulum (m) & $1$	\\
			$g$ & Gravitational const. (m/s$^2$) & $9.8$	\\
			$T_s$ & Sampling period (s) & $0.05$	\\
			$\vx_0$ & Initial state & $[\pi, 0]^{\top}$ \\
			$\vmu_w$ & Mean of disturbance & $[0, 0.5]^{\top}$ \\
			$\mSigma_w$ & 	Variance-covariance matrix of disturbance & $\mathrm{diag}(0.05^2, 0.1^2)$ \\
			$\eta$ & Lower bound of probability of constraint satisfaction 	& $0.95$ \\
			$\xi$ & Lower bound of probability coming back to $\gX_s$	& $0.9998$ \\
			$\tau$ & Maximum steps you need to get back to $\gX_s$		& $2$ \\
			$\angv^{\max}(-\angv^{\min})$ & Upper and lower bounds of angular velocity (rad/s)& $6.0$  \\
        \hline
		\end{tabular}
\end{table}

\begin{table}[htbp]
	\caption{Simulation parameters (Four-bar parallel link robot manipulator)}
	\label{tb:sim_para_Mani}
	\centering
		\begin{tabular}{|l|l|l|}
		\hline
		Symbol  &Definition &Value \\
		\hline
			$T$ & \begin{tabular}{l}Number of simulation steps\end{tabular} & $100$	\\
			$N$ & \begin{tabular}{l}Number of learning episodes\end{tabular} & $100$	\\
			$\hat{m}_{11}$ & \begin{tabular}{l}Model parameter\end{tabular} & $3.91\times 10^{-3}$ \\
			$\hat{m}_{22}$ & \begin{tabular}{l}Model parameter\end{tabular} & $2.39 \times 10^{-3}$ \\
			$\hat{d}_{11},\ \hat{d}_{22}$ & \begin{tabular}{l}Model parameter\end{tabular} & $9.37\times 10^{-3} $ \\
			$V_{1}$ & \begin{tabular}{l}Model parameter\end{tabular} & $9.01 \times 10^{-2} $ \\
			$V_{2}$ & \begin{tabular}{l}Model parameter\end{tabular} & $1.92 \times 10^{-2} $ \\
			$\alpha$ & \begin{tabular}{l}Model parameter\end{tabular} & $6.89\times 10^{-2}$ \\
			$T_s$ & \begin{tabular}{l}Sampling period (s)\end{tabular} & $0.05$	\\
			$\vx_0$ & \begin{tabular}{l}Initial state\end{tabular} & $[\pi, \pi,0, 0]^{\top}$ \\
			$\vmu_w$ & \begin{tabular}{l}Mean of disturbance\end{tabular} & $[0,0.1,-0.1, 0.05]^{\top}$ \\
			$\mSigma_w$ & 	\begin{tabular}{l}Variance-covariance matrix \\ \ \ of disturbance \end{tabular}& 
			        $ \mathrm{diag}( 0.01^2, 0.03^2, 0.02^2,0.01^2)$  \\
			$\eta$ &\begin{tabular}{l}Lower bound of probability of \\ \ \ constraint satisfaction\end{tabular}  	& $0.95$ \\
			$\xi$ & \begin{tabular}{l}Lower bound of probability \\ \ \ coming back to $\gX_s$	\end{tabular} & $0.9998$ \\
			$\tau$ & \begin{tabular}{l}Maximum steps you need to \\ \ \ get back to $\gX_s$\end{tabular}		& $2$ \\
			$\varpi^{\max}(-\varpi^{\min})$ & \begin{tabular}{l}Upper and lower bounds of \\ \ \ rotating speed (rad/s)\end{tabular}& $6.0$  \\
		\hline
		\end{tabular}
\end{table}

\begin{table}[htbp]
	\caption{Hyperparameters of DDPG algorithm using Adam}
	\label{tb:sim_para_ddpg}
	\centering
		\begin{tabular}{|l|l|l|}
		\hline
		Symbol  &Definition &Value \\
		\hline
			$\iota_a$ & \begin{tabular}{l}Learning rates for actor network\end{tabular}& $1.0\times10^{-3}$ \\
			$\iota_c$ & \begin{tabular}{l}Learning rates for critic network\end{tabular}& $2.0\times10^{-3}$  \\
			n/a & \begin{tabular}{l}Weight decay\end{tabular}& $0$  \\
			$\gamma$ & \begin{tabular}{l}Discount factor\end{tabular} & $0.99$ \\
			$\tau^{\mathrm{DDPG}}$ & \begin{tabular}{l}Factor for the soft targets updates\end{tabular}& $5.0\times10^{-3}$  \\
			n/a & \begin{tabular}{l}Capacity of replay buffer\end{tabular}& $5.0\times10^{5}$  \\
			n/a & \begin{tabular}{l}Minibatch size\end{tabular}& $64$  \\
		\hline
		\end{tabular}
\end{table}
